\documentclass{article} 
\usepackage{iclr2026_conference,times}
\iclrfinalcopy

\usepackage{amsmath,amsfonts,bm}

\def\eqref#1{equation~\ref{#1}}

\def\1{\bm{1}}

\DeclareMathAlphabet{\mathsfit}{\encodingdefault}{\sfdefault}{m}{sl}
\SetMathAlphabet{\mathsfit}{bold}{\encodingdefault}{\sfdefault}{bx}{n}

\usepackage[
    colorlinks=true,
    citecolor=blue,
    linkcolor=purple,
    urlcolor=black
]{hyperref}
\usepackage{url}
\usepackage{booktabs}       
\usepackage{amsmath,amsfonts,amssymb}       
\usepackage{nicefrac}       
\usepackage{microtype}      
\usepackage{xcolor}         
\usepackage{float}

\usepackage{array}    

\usepackage{algorithm,algorithmic}
\usepackage{adjustbox}

\usepackage{wrapfig}

\usepackage{graphicx}
\usepackage{subcaption}
\usepackage{booktabs} 
\usepackage{chngcntr}
\usepackage{tabularx, colortbl}
\usepackage{multirow}

\usepackage{enumitem}

\usepackage{amsmath}
\usepackage{amssymb}
\usepackage{mathtools}
\usepackage{amsthm}

\newcommand\PLM{P_{\!\textsc{lm}}}
\usepackage{newtxmath}

\makeatletter

\renewcommand{\appendixautorefname}{\S\@gobble}
\renewcommand{\sectionautorefname}{\S\@gobble}
\renewcommand{\subsectionautorefname}{\S\@gobble}
\renewcommand{\subsubsectionautorefname}{\S\@gobble}

\makeatother

\usepackage{listings}         
\lstdefinelanguage{text}{
   alsoletter={}
}
\makeatletter

\renewcommand{\paragraph}[1]{\textbf{#1}}
\providecommand{\subparagraph}{}
\renewcommand{\subparagraph}{%
  \@startsection{subparagraph}{5}{\z@}%
                {1.5ex \@plus 0.5ex \@minus 0.2ex}%
                {-1em}%
                {\normalsize\bf}%
}

\title{Latent Veracity Inference for Identifying\\Errors in Stepwise Reasoning}

\author{Minsu Kim\textsuperscript{1, 2 *} \quad Jean-Pierre Falet\textsuperscript{1, 3, 4 *} \quad  Oliver E. Richardson\textsuperscript{1, 3} \quad Xiaoyin Chen\textsuperscript{1, 3} \\ \bf  Moksh Jain\textsuperscript{1, 3} \quad
Sungjin Ahn\textsuperscript{2} \quad Sungsoo Ahn\textsuperscript{2} \quad Yoshua Bengio\textsuperscript{1, 3, 4}\vphantom{\thanks{Equal contribution. }}\\
    \textsuperscript{1}Mila -- Qu\'ebec AI Institute
    \quad\textsuperscript{2}KAIST
    \quad\textsuperscript{3}Universit\'e de Montr\'eal
    \quad\textsuperscript{4}LawZero
    \\
\small \texttt{\{minsu.kim, jean-pierre.falet\}@mila.quebec}
}

\begin{document}

\maketitle

\begin{abstract}
Chain-of-Thought (CoT) reasoning has advanced the capabilities and transparency of language models (LMs); however, reasoning chains can contain inaccurate statements that reduce performance and trustworthiness. 
To address this, we propose to augment each reasoning step in a CoT with a latent veracity (or correctness) variable.
To efficiently explore this expanded space, we introduce \emph{Veracity Search} (VS), a discrete search algorithm over veracity assignments. It performs otherwise intractable inference in the posterior distribution over latent veracity values by leveraging the LM's joint likelihood over veracity and the final answer as a proxy reward.
This efficient inference-time verification method facilitates supervised fine-tuning of an \emph{Amortized Veracity Inference} (AVI) machine by providing pseudo-labels for veracity. AVI generalizes VS, enabling accurate zero-shot veracity inference in novel contexts. Empirical results demonstrate that VS reliably identifies errors in logical~(\textsc{ProntoQA}), mathematical~(\textsc{GSM8K}), and commonsense~(\textsc{CommonsenseQA}) reasoning benchmarks, with AVI achieving comparable zero-shot accuracy. Finally, we demonstrate the utility of latent veracity inference for providing feedback during self-correction and self-improvement.
\end{abstract}

\section{Introduction} \label{sec:intro}

The inference-time compute paradigm---the practice of allowing language models (LMs) to generate a chain-of-thought (CoT) before producing an answer---has lead to a improvements in reasoning performance across a variety of domains~\citep{nye2021show,kojima2022large,wei2022chain}.
The CoTs themselves also promise a degree of interpretability, giving operators a tool to detect problematic behavior~\citep{perez2023discovering}.
This promise, however, is undermined by the fact that LMs often generate flawed reasoning steps~\citep{ji2023survey, zhang2023language, bang2023multitask}.
Flawed reasoning impairs interpretability and may propagate to the model’s final output~\citep{cobbe2021training,zelikman2022star,wang2023selfconsistency,yao2023tree,turpin2023language,lightman2024lets}, making error detection (and correction) an important challenge for improving LM trustworthiness.

Several methods have been proposed to improve the correctness, or \emph{veracity}, of a CoT. Training on labeled reasoning steps is one solution~\citep{camburu2018esnli,rajani-etal-2019-explain,lightman2024lets}, but it is impeded by the paucity of comprehensive annotated datasets due to high labeling cost. Fact verification through evidence retrieval from external corpora represents a compromise in terms of labeling requirements~\citep{chern2023factool, min2023factscore, jacovi2024chain}, but faces other obstacles including retrieval complexity and evidence coverage gaps.

We propose a new method to automatically identify stepwise errors in a CoT without requiring supervision for each reasoning step.
Our key idea is to formulate the problem of identifying errors as the problem of doing posterior inference in a latent-variable model (LVM) where each reasoning step is augmented with a latent veracity variable indicating its correctness. This label can be binary (True or False) for many applications, but our framework is also compatible with categorical variables that can take on more than two values. The CoT itself, along with the final output of the reasoning process (the answer to a query), are treated as observations that serve as the main signal for inferring accurate latent-veracity assignments. 
In more detail, our contributions are as follows:
\begin{description}[itemsep=0pt, 
parsep=2pt,topsep=0pt,
    leftmargin=1.5em,
    ]
    \item[Section~\ref{sec:lvm}:]We cast stepwise error identification as a latent-variable modeling problem. 
    \item[Section~\ref{sec:search-correction}:]We introduce a discrete search algorithm, Veracity Search (VS), which leverages the LM’s joint likelihood over stepwise veracity and the final answer as a proxy reward for approximately sampling from the the target distribution over latent veracity assignments, highlighting differences with standard methods that use in-context learning to turn LMs into verifiers. 
    \item[Section~\ref{sec:amortized_corrector}:]We propose Amortized Veracity Inference (AVI) to train an LM to predict a distribution over veracity assignments that does not depend on the true answer, using supervised fine-tuning on pseudo-labels obtained from VS. As a result, AVI enables zero-shot latent-veracity inference in downstream reasoning tasks where the final answer is unknown.
    \item[Section~\ref{sec:exp}:]We validate our approach on the logical reasoning benchmark \textsc{ProntoQA}~\citep{saparov2022language}, the mathematical reasoning task \textsc{GSM8K}~\citep{cobbe2021training}, and commonsense reasoning \textsc{CommonsenseQA}~\citep{talmor2018commonsenseqa} using several open-source LMs (Qwen~\citep{bai2023qwen, qwen2, qwen2.5}, Llama~\citep{touvron2023llama, touvron2023llama2, grattafiori2024llama}). Our method yields consistent improvements in verification accuracy over in-context learning baselines, and scales to longer and more complex reasoning chains. We demonstrate the utility of error-identification in downstream tasks by using veracity assignments as feedback for self-correction and self-improvement~\citep{pan2023automatically}.
\end{description}

\section{Method}
\label{sec:method}

Let $\PLM$ denote an LM's probability distribution over the set of possible sequences of tokens. Modern ``thinking'' LMs process an input prompt $x$ by first generating a CoT $z$ and subsequently producing the final answer $y$.
Marginalizing out the CoT, the model's distribution of outputs given the input $x$ is 
$
    \mathbb P(y \mid x)
    = \sum_{z} \PLM(y\, z \mid x)
    = \mathbb{E}_{z \sim \PLM(z \mid x)}\left[ \PLM(y \mid x\, z) \right].
$
(Note that we use a different symbol, $\mathbb P$, to distinguish this probabilistic model from the result $\PLM(y\mid x)$ of directly querying the LM.)
Some approaches (e.g., \citet{zelikman2022star, hu2023amortizing}) replace $\PLM(z\mid x)$ with a learned distribution~$Q(z \mid x)$ that puts more weight on ``correct'' reasoning chains by training $Q$ to approximate the distribution $\mathbb P(z \mid x, y^*)$, where $y^*$ is the true answer to the query. Underpinning these approaches is the assumption that $\mathbb P(y^* \mid x)$ is increased as a result of marginalizing with respect to such a $Q$.
Typical strategies for training~$Q$ include  
(i) supervised fine-tuning on labeled examples of $(x,z,y)$~\citep{gulcehre2023reinforced}, and 
(ii) reinforcement learning (e.g., REINFORCE~\citep{zelikman2022star}), amortized inference (e.g., GFlowNets~\citep{hu2023amortizing}), or test-time inference (e.g., Sequential Monte Carlo~\citep{pmlr-v235-zhao24c}) using $\PLM(z \, y ^* \mid x)$ as a reward, to approximately sample from the intractable posterior $\mathbb P(z \mid x, y^*)$.
An analogue of this distribution $Q$ lies at the heart of our proposal, but to explain it, we must first formally introduce an additional dimension: veracity.

\subsection{A Latent‐Variable Model (\texorpdfstring{$\mathbb P$}{P}) Augmented With Veracity (\texorpdfstring{$V_z$}{Vz})}
\label{sec:lvm}

One should not expect every reasoning step in a CoT to be correct.
Yet, by viewing $z$ as a logical statement and conditioning on it, standard practice implicitly identifies $z$ with the proposition that ``$z$ is correct''. 
    From this starting point, it is clear why so much work has gone into correcting reasoning chains: so doing would validate the unstated assumption.
Our approach is different: we take the \textit{identity} of the CoT $z$ as given and introduce a new binary variable $V_z$ intended to capture the \textit{veracity} of $z$.
Since the standard assumption that $V_z$ is always $1$ does not always hold in practice, we claim that identifying incorrect explanations $z$ is crucial.

As commonly done in stepwise CoT evaluation~\citep{golovneva2022roscoe, lightman2024lets, manakul2023selfcheckgpt}, we parse the CoT $z$ into a sequence of primitive statements $z = (z_1, z_2, \ldots, z_N)$; as a result, boolean veracity $V_z$ is a binary random vector taking values in $\{0,1\}^N$.
We imagine that the distribution over $V_z$ and $Y$ given $x$ and $z$ is governed by the behavior of an auto-regressive LM, which gives us a (conditioned) LVM $\mathbb P$ over the variables $V_z$ and $Y$: 
\begin{equation}
    \mathbb{P}(V_z{=}v, Y{=}y \mid x,z)
        \;:=\; \PLM(v\, y \mid x\,z)
        \;=\; \PLM(v \mid x\,z)\, \PLM(y \mid x\,z\,v).
\end{equation}
The veracity variables $V_z$, however, are not observed, so calculating $\mathbb P(V_z \mid x,z,y)$ requires summing over the $2^N$ possible values of $V_z$ to calculate the denominator $\mathbb P(Y\mid x,z)$:
\begin{equation}
\label{eq:bayes_rule}
\begin{aligned}
     \mathbb{P}(V_z{=}v \mid Y{=}y, x, z) &= \frac{\mathbb P (V_z{=}v, Y{=}y \mid x\, z)}{\mathbb P (Y=y \mid x\, z)}=\frac{\PLM(v \mid x\, z) \PLM(y \mid x\, z\, v)}{\sum_{v' \in \{0,1\}^N} \PLM(v' \mid x\, z) \PLM(y \mid x\, z\, v')},
\end{aligned}
\end{equation}
From another angle, one might view the problem as about the fixed-order auto-regressive nature of the LM, which makes joint probabilities under $\PLM$ sensitive to the order of the sequence. If we assume the joint factorization in Eq.~\ref{eq:bayes_rule}, corresponding to the order one would expect to sample veracity in if it is to be useful for predicting $Y$, then the problem of doing inference in a conditional LVM $\mathbb P$ in which $Y$ is observed but $V_z$ is not amounts to infilling
$V_z$ in the sequence $X \rightarrow Z \rightarrow (V_z) \rightarrow Y$.
For this reason, $\mathbb{P}(V_z{=}v \mid Y{=}y, x, z)$ is likely to differ from $\PLM(v \mid x\,z\,y)$; the latter corresponds to an alternative generative model $X \to Z \to Y \to V_Z$, and may not be a good approximation the posterior $\mathbb P(V_z \mid x,z, y)$ of interest in our LVM.
Nevertheless, $\PLM(V_z{=}v \mid x\,z\,y)$ corresponds to an obvious in-context learning baseline where the LM is prompted to predict the veracity of a reasoning chain $z$. We evaluate against such baselines in Section \ref{sec:exp}.

\begin{figure*}[t]
  \centering
  \vspace{-0.5em}
  \includegraphics[width=0.9\linewidth]{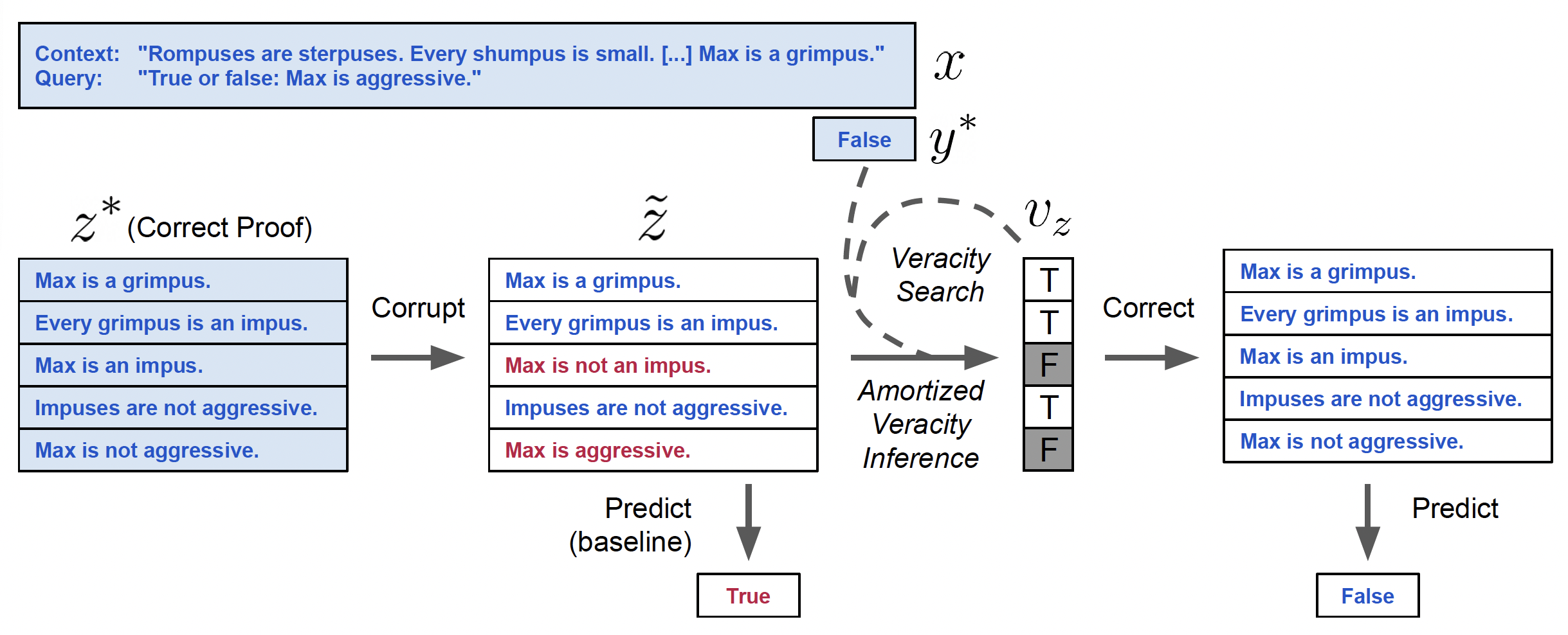}
  \vspace{-1em}
  \caption{\textbf{Overview of our latent veracity inference method applied to \textsc{ProntoQA}.} Given an input $x$, the \emph{Veracity Search} (VS) takes an erroneous CoT $\tilde{z}$ and searches for a veracity vector $v_z$ with high joint likelihood $\PLM(v_z\; y^* \mid x \, z)$, where $y^*$ is the correct answer. Veracity vectors can then be used as pseudo-labels to fine tune an LM via \emph{Amortized Veracity Inference} (AVI) for zero-shot veracity inference, eliminating the dependencies on $y^*$ and on the test-time search algorithm (dotted lines). A LM can use veracity assignments for correcting flawed reasoning steps.}
  \label{fig:figure1}
  \vspace{-1em}

\end{figure*}

Tying this back to the related work referenced at the top of \S\ref{sec:method}, we therefore seek a variational posterior distribution~$Q(V_z\mid x,z)$ over veracity values that puts more weight on ``correct'' veracity assignments by training it to approximate the intractable posterior $\mathbb P(V_z \mid x,z,y^*)$. 
The difficulty of doing so motivates us to design an efficient search-based approach that performs iterative refinement of $V_z$ (VS; \S\ref{sec:search-correction}), which will ultimately be the stepping stone to get such a $Q$ using AVI (\S\ref{sec:amortized_corrector}). An overview of our approach in the context of a logical reasoning task is summarized in Fig.~\ref{fig:figure1}. 

\subsection{Veracity Search (VS)}
    \label{sec:search-correction}

Assume that the query $x$, CoT $z$ (possibly with errors), and---for now---the correct answer $y^*$ are given.
Define the \emph{proxy reward} for a bit vector $v \in \{0,1\}^N$ as
\begin{equation}
\label{eq:reward}
\begin{aligned}
    R(v)
    \;:=\;
    \mathbb{P}(V_z{=}v, Y{=}y^* \mid x, z) 
       &= \PLM(v\,y^* \mid x \, z) \propto \mathbb{P}(V_z{=}v \mid Y{=}y^*, x, z), 
\end{aligned}
\end{equation}
where the proportionality relation is obtained via the application of Bayes rule in Eq.~\ref{eq:bayes_rule}. VS seeks high-reward assignments  
$v_z\!\in\!\{0,1\}^{N}$,
which corresponds to sampling from $\mathbb P(V_z \mid x,z, y^*)$, the latent variable model's posterior over veracity assignments. 
In comparison to methods that produce better reasoning chains using a reward signal coming only from the target answer $Y$ \citep{cobbe2021training, zelikman2022star}, or that treats the CoT as a latent variable $Z$ that entangles identity and veracity~\citep{hu2023amortizing, phan2023training}, our proxy reward is taken from a latent variable model relating both veracity and final answers, and enables fixing the identity of the CoT $Z=z$ and focusing on the sub-problem of veracity inference. 
In \S~\ref{sec:amortized_corrector}, we will see a way of overcoming the requirement for the true label $y^*$, which is required for downstream reasoning tasks: by training an amortized veracity sampler $Q(V_Z \mid x, z)\propto R(V_z)$ (or a low-temperature variant for approximate maximization).

\paragraph{Working Hypothesis.} While LMs often struggle to generate logically sound and consistent CoTs during sampling, we hypothesize they are nevertheless capable of assigning higher probability to the joint distribution over the true answer and the veracity of a reasoning chain when the latter is closer to the ground-truth $v_z^*$. More formally, this means that we expect a negative correlation between
$
\mathbb P(V_z =v_z \mid Y={y^*}, x, z)
$
and the Hamming (L1) distance $|v_z-v_z^*|$.

While the global maximizer of the likelihood may not always coincide exactly with $v_z^*$, moving towards higher likelihoods should, on average, steer $v_z$ toward the true assignment $v_z^*$ and thereby reveal which statements in~$z$ are correct. We empirically validate this hypothesis in Appendix~\ref{app:correlation}.

\paragraph{Single-Bit Metropolis Updates with Simulated Annealing.}
At iteration $t \in \{1, 2, \ldots,\}$ we take the current vector $v_z^{(t)}$ and perform a
\emph{single-coordinate} update:
\begin{enumerate}[noitemsep,topsep=0pt,
    leftmargin=1cm,rightmargin=0.7cm
    ]
    \item Draw an index $j \sim \mathrm{Unif}\{1,\dots,N\}$, and propose the veracity vector obtained by flipping the $j$-th bit: $v'_z = v_z^{(t)} \oplus e_j$ (e.g.\ if $j=3$ and $v_z^{(t)}=[1,0,1]$, then $v'_z=[1,0,\mathbf{0}]$).
    \item Accept the proposal $v'_z$ with probability
          $
                \alpha_t
                \;=\;
                \min\Bigl\{1,\;
                    \bigl[R(v'_z)/R(v_z^{(t)})\bigr]^{\beta_t}
                \Bigr\},
            $
         where inverse temperature $\beta_t$ is set to 1 in the basic Metropolis algorithm. We also experiment with a schedule for $\beta_t$, $0<\beta_0<\dots<\beta_{M}$, in the style of simulated annealing, to balance the exploration and exploitation tradeoff. 
    \item Set $v_z^{(t+1)} = v'_z$ if accepted, otherwise let $v_z^{(t+1)}=v_z^{(t)}$.
\end{enumerate}
Because the proposal flips exactly one bit and is symmetric,
the acceptance ratio simplifies to the likelihood ratio. 

\paragraph{Greedy–Tree Initialization.} \label{sec:greedy-search}
Before running the stochastic search, we heuristically pick an initial
vector $v_z^{(0)}$ using a depth-first greedy procedure.
For $i=1$ to $N$:
\begin{enumerate}
    [leftmargin=1cm,rightmargin=0.7cm]
    \item Construct partial candidates that agree on positions $1{:}i{-}1$
          and set $v_{z_i}\!=\!0$ or $1$, leaving $i{+}1{:}N$ unassigned.
    \item Query the model for the partial score
          $\tilde R(v_{1:i})=\PLM\bigl(x\; z_{1:i}\; v_{1:i}\; y^* \bigr)$,
          letting the LM internally marginalize the unspecified bits.
    \item Fix $v_{z_i}^{(0)}$ to the value that yields the larger $\tilde R$.
\end{enumerate}
This greedy sequential ``tree'' search provides a warm start that is
often close to a high-reward basin, allowing the subsequent  
single-bit Metropolis updates to converge more quickly.

\subsection{Amortized Veracity Inference (AVI)}
\label{sec:amortized_corrector}

In the spirit of variational expectation–maximization (EM), higher-reward samples of the veracity vector $V_z$ obtained via VS (Eq.~\ref{eq:reward}) are used as pseudo-labels to train the generative model $\PLM$ via supervised fine-tuning, yielding the amortized sampler $Q$. Initializing at $\PLM(V_z \mid x\, z)$ and fine-tuning on VS samples obtained with simulated annealing towards zero temperature results in a an approximate maximizer of the proxy-reward $Q(V_z \mid x, z) \propto \lim_{\beta \rightarrow \infty }R(V_z)^{\beta}$.

Unlike the test-time search method (VS), $Q(V_z \mid x, z)$ is trained to predict veracity without conditioning on the answer $Y$. This provides two primary benefits: (1) it enables rapid run-time inference for verifying a CoT $z$ in a zero-shot manner without test-time search and before seeing the correct answer, and (2) it has the potential to improve reasoning by serving as feedback to identify erroneous reasoning steps in need of correction or re-generation. In some tasks (like \textsc{ProntoQA}), reasoning steps predicted to be incorrect by AVI can be negated (by inserting a ``not'' token), yielding a corrected explanation $z'$. The correction is expected to be a more accurate reasoning chain, i.e. $\PLM(y^* \mid x\, z') > \PLM(y^* \mid x\,z)$, and we empirically validate this hypothesis in \S~\ref{sec:exp_amortized}. For tasks where negation can't be used for correction, a more general strategy involves simply appending the predicted veracity label (``True'' or ``False'') to the statement, and we showcase this approach in our \textsc{CommonsenseQA} experiments.

\section{Related Work}
\label{sec:related_work}

\paragraph{Automated feedback, self-correction, and self-improvement.} 
Several self-correction and self-improvement methods~\citep{pan2023automatically} rely on critics to provide feedback to a reasoning model, with the aim of correcting a CoT or guiding its re-generation. Sources of feedback include supervised classifiers and reward models~\citep{rajani-etal-2019-explain, lightman2024lets, cobbe2021training, yang-etal-2022-generating}, evidence-based fact verifiers~\citep{chern2023factool, li2023self, manakul2023selfcheckgpt}, and zero/few-shot prompted LMs (to provide natural language feedback, or scalar scores)~\citep{madaan2023self, weng2022large, xie2023self, yao2023tree, shinn2023reflexion}. At training time, the final-answer itself can be used to fine-tune a reasoning model to output reasoning chains that reach the correct conclusion~\citep{zelikman2022star}, which may involve variational EM to  learn---and do inference in---an LVM relating reasoning chains and final answers~\cite{phan2023training, hu2023amortizing}. Iterative approaches based on in-context learning to teach LMs to correct rationales by contrasting them with examples of correct rationals have also been proposed~\citet{zhou2024can}.
RL for eliciting CoT reasoning, specifically with binary feedback based on the answers produced by CoTs, has become prominent following models like o1~\citep{openai2024o1} and DeepSeek-R1~\citep{guo2025deepseek}.

Our work can be viewed as \emph{complementary} to self-correction and self-improvement methods, proposing a new method for identifying errors that can serve as a useful source of feedback for any self-correction/self-improvement method; alternatively, it can be useful on its own for monitoring purposes. In contrast to standard feedback obtained via zero- or few-shot prompting of an LM, which has been found to be fragile (in part due to prompt sensitivity) and degrade performance in certain cases~\citep{huang2022large}, our proposed method samples directly from the posterior distribution of interest in an LVM relating stepwise veracity to final answers. It does not require examples of correct reasoning chains, and the required instruction-prompt is minimal. Decomposing self-correction into verification and refinement sub-tasks has also been proposed in~\citet{Zhang2024SmallLM}, outside the context of LVMs.

\paragraph{Process reward models (PRMs).} PRMs score the quality of intermediate reasoning steps and are widely used in LM post‑training. They differ from our method in two key ways: (i) typically, PRMs require step-level ground-truth labels (e.g. provided by humans~\citep{lightman2024lets}); (ii) supervised/reinforcement-learning fine-tuning methods to train PRMs using only outcome-supervision (final‑answer labels) learn stepwise rewards that represent value/advantage and capture instrumental utility rather than stepwise correctness, and can reward useful‑but‑incorrect steps~\citep{uesato2022solving, zha2025rl, yuan2025free, cui2025process}. In contrast, VS and AVI explicitly target step \textit{veracity} without the need for process supervision, by approximating latent-veracity inference in an LVM.

\paragraph{Search-based inference.} Search-based inference complements prompt engineering and fine-tuning to enhance reasoning in LMs. Simple approaches utilize CoT resampling and majority-voting to achieve improvements over single-pass generation~\citep{wang2023selfconsistency, xue2023dynamic}. \emph{Best-of-N} methods extend the sampled candidate pool and re-ranks outputs via specialized ranking models~\citep{cobbe2021training, snell2024scaling}; however, these methods scale linearly with the number of candidates and do not revise intermediate reasoning steps. Others frame reasoning as a combinatorial search problem, exemplified by tree search prompting, which explores and prunes reasoning branches through learned or heuristic value functions~\citep{yao2023tree, xie2023self}, its extensions involving differentiable relaxations~\citep{xu2025softcot}, amortized inference with GFlowNets~\citep{bengio2021flow, yu2024flow}, and Monte-Carlo Tree Search~\citep{luo2024improve, zhang2024accessing, xie2024monte}. Our VS method is based on local search (MCMC), which is efficient in the lower-dimensional search space of veracity, and it borrows several ideas from the related work cited above.

\section{Experiments}
\label{sec:exp}

In this section we evaluate VS emperically,\footnote{Source code: \url{https://github.com/alstn12088/veracity_inference}}  validating our hypothesis that using $\PLM(v_z \;y^* \mid x\, z)$ as a proxy reward improves the correctness of the veracity vector (with respect to the ground-truth $v^{*}_z$).

\paragraph{Benchmarks.}  First, we use the \textsc{ProntoQA} benchmark~\citep{saparov2022language}, because it gives us the ability to synthesize correct reasoning chains ($z^*$) synthetically, facilitating the introduction of controlled errors by corrupting specific steps, resulting in an incorrect proof $\tilde{z}$. Key advantages of \textsc{ProntoQA} include binary veracity labels for each logical deduction steps, and the ability to generate proofs from a fictional ontology to isolate the LM's logical reasoning capability from it's ability to retrieve memorized facts/trends acquired during training. Moreover, \textsc{ProntoQA} allows for adjusting the reasoning chain length to test our method's scalability to more complex scenarios. Finally, we use \textsc{ProntoQA} to assess AVI and its impact on answer prediction ($y^*$).

We additionally evaluate our method across other reasoning domains, namely mathematical reasoning~\citep[\textsc{GSM8K}]{cobbe2021training} and commonsense reasoning~\citep[\textsc{CommonsenseQA}]{talmor2018commonsenseqa}. In contrast to \textsc{ProntoQA}, these datasets do not provide us with the ability to generate corrupted reasoning chains in a controlled manner, limiting our ability to measure the accuracy of veracity assignments. We overcome this by generating structured reasoning chains via a more powerful oracle model (\texttt{GPT-4.1}) using OpenAI’s API for generating structured outputs (a sequence of strings, each of which corresponds to a reasoning steps), conditioned on the correct answer $y^*$. We treat the generated reasoning chains as ground truth, assuming $v_z^*=1$, and then corrupt them with controlled perturbations.

\paragraph{Base LMs.} 
We evaluate our approach using several representative LMs: Qwen 3 (4B), Qwen 3 (8B), Llama 3.2 (3B), and Llama 3 (8B).

\paragraph{Baselines.} 
Our primary goal is to demonstrate the effectiveness of using the joint probability $\PLM(v_z\, y^* \mid x\, z)$ as a proxy reward. We introduce autoregressive baselines that generate $V_z$ in a tractable manner by modifying the decoding trajectory from the original intractable form $(X \rightarrow Z \rightarrow V_z \rightarrow Y)$ to the tractable form $(X \rightarrow Y \rightarrow Z \rightarrow V_z)$. These baseline methods directly query the LM to generate $V_z$ given $x$, $y$, and $z$, using few-shot examples.

These baseline inference methods are further categorized as follows: (1) Many2Many Inference (\textbf{Many2Many}): Generates a complete veracity vector $v_z$ in one-shot for a given sequence of reasoning steps $z$; (2) Many2Many Inference with the addition of CoT-prompting (\textbf{CoT}): Given $x$, $y$, and $z$, we prompt the model to generate a CoT before outputting the value of $V_{z}$, to explain its reasoning; (3) One-Shot Majority Voting (\textbf{Voting}): Uses majority voting across $M=50$ samples generated at a higher temperature ($T=0.5$) to predict the veracity assignment, whereas other methods employ a greedy-like temperature ($T=0.01$) to enhance consistency; (4) Recursive Inference (\textbf{Recursive}): Predicts each stepwise veracity label $v_{z_i}$ recursively, where $i$ corresponds to the index of the statement $i$ in the reasoning chain, conditioned on previously inferred labels $v_{z_{1:i-1}}$ and corresponding statement identities $z_{1:i-1}$. The generative trajectory is thus structured as $X \rightarrow Y \rightarrow Z_1 \rightarrow V_{z_1} \rightarrow \cdots \rightarrow Z_N \rightarrow V_{z_N}$. We provide the same five few-shot demonstrations for all baselines, including our proposed method.

Details pertaining to implementation, CoT corruption, and prompting, are provided in Appendix~\ref{app:implementation}.

\begin{table}[t]
  \caption{Mean Hamming Similarity ($\pm$ std) on \textsc{ProntoQA}, \textsc{GSM8K}, and \textsc{CommonsenseQA} (1,000 examples each).}
  \label{tab:combined_scores}
  \centering
  \small
  \resizebox{1.0\linewidth}{!}{
  \begin{tabular}{llcccc}
    \toprule
    Dataset & Method & Qwen-4B & Qwen-8B & Llama-3B & Llama-8B \\
    \midrule
    \multirow{5}{*}{\textsc{ProntoQA}}
      & Recursive        & $0.691 \pm 0.167$ & $0.667 \pm 0.139$ & $0.538 \pm 0.117$ & $0.471 \pm 0.057$ \\
      & Many2Many        & $0.590 \pm 0.155$ & $0.683 \pm 0.142$ & $0.506 \pm 0.161$ & $0.530 \pm 0.157$ \\
      & Voting           & $0.603 \pm 0.152$ & $0.692 \pm 0.138$ & $0.514 \pm 0.156$ & $0.536 \pm 0.153$ \\
      & CoT              & $0.591 \pm 0.201$ & $0.384 \pm 0.201$ & $0.459 \pm 0.041$ & $0.515 \pm 0.162$ \\
      & VS (ours) & $\mathbf{0.910} \pm 0.118$ & $\mathbf{0.945} \pm 0.096$ & $\mathbf{0.948} \pm 0.072$ & $\mathbf{0.964} \pm 0.072$ \\
    \midrule
    \multirow{5}{*}{\textsc{GSM8K}}
      & Recursive        & $0.540 \pm 0.167$ & $0.617 \pm 0.136$ & $0.568 \pm 0.096$ & $0.568 \pm 0.096$ \\
      & Many2Many        & $0.620 \pm 0.126$ & $0.650 \pm 0.139$ & $0.566 \pm 0.096$ & $0.567 \pm 0.096$ \\
      & Voting           & $0.623 \pm 0.127$ & $0.654 \pm 0.138$ & $0.566 \pm 0.096$ & $0.553 \pm 0.128$ \\
      & CoT              & $0.614 \pm 0.166$ & $0.695 \pm 0.204$ & $0.496 \pm 0.164$ & $0.496 \pm 0.165$ \\
      & VS (ours) & $\mathbf{0.711} \pm 0.155$ & $\mathbf{0.751} \pm 0.193$ & $\mathbf{0.614} \pm 0.143$ & $\mathbf{0.646} \pm 0.157$ \\
    \midrule
    \multirow{5}{*}{\textsc{CommonsenseQA}}
      & Recursive        & $0.607 \pm 0.217$  & $0.509 \pm 0.220$  & $0.505 \pm 0.219$& $0.506 \pm 0.219$  \\
      & Many2Many        & $0.517 \pm 0.227$ & $0.534 \pm 0.212$ & $0.504 \pm 0.220$ & $0.503 \pm 0.219$  \\
      & Voting           &$0.521 \pm 0.226$  & $0.533 \pm 0.208$ & $0.504 \pm 0.220$ & $0.505 \pm 0.219$ \\
      & CoT              & $0.695 \pm 0.230$ & $0.590 \pm 0.220$ & $0.507 \pm 0.219$ & $0.535 \pm 0.227$ \\
      & VS (ours) & $\mathbf{0.935} \pm 0.123$ & $\mathbf{0.931} \pm 0.119$ & $\mathbf{0.836} \pm 0.176$ & $\mathbf{0.903} \pm 0.137$  \\
    \bottomrule
  \end{tabular}}
\end{table}
\subsection{Evaluating the Accuracy of Veracity Inference}

To quantify the performance of VS, we compute the Hamming similarity between the predicted correctness vector \(v_{z}\) and the ground-truth vector \(v^{*}_{z}\): $\operatorname{Sim}(v_z,v_z^*)
\;=\;
1 \;-\;
\lVert v_z - v_z^*\rVert_{1}/{L}$,
where \(\lVert\cdot\rVert_{1}\) denotes the element-wise \(\ell_{1}\)-norm and \(L = \lvert v_z\rvert = \lvert v_z^*\rvert\) is the length of the vector. A similarity of~1 indicates perfect agreement, whereas~0 signifies complete disagreement. 

The results for \textsc{ProntoQA}, \textsc{GSM8K}, and \textsc{CommonsenseQA} are presented in Table~\ref{tab:combined_scores}. In both tasks, VS (run for 200 iterations on the proxy-reward Eq.~\ref{eq:reward}), consistently outperforms the other baselines.  In particular, VS attains near-perfect identification accuracy on \textsc{ProntoQA}.

In the \textsc{GSM8K} mathematical reasoning task, VS consistently outperformed baseline methods across all tested LMs. The remaining gap with ground-truth veracity labels is likely due to the distribution of errors induced by our approach of corrupting CoTs with noise.
In cases where a reasoning step can be viewed as a Boolean assertion (\textsc{ProntoQA} and \textsc{CommonsenseQA}), flagging errors is no different from correcting them. In the setting of \textsc{GSM8K}, however, beyond flagging arithmetic errors, one also needs to re-do the flawed calculations in order to correct the answer or identify additional errors, impacting both the baselines and VS. 

The importance of the CoT corruption scheme also highlights the need to evaluate VS under other distributions of erroneous reasoning chains. 
We conduct such an evaluation on \textsc{ProntoQA} and \textsc{CommonsenseQA} in Appendix~\ref{app:different_patterns}, and find that VS outperforms baselines in a wider variety of settings.
Finally, we evaluate VS in cases where veracity is a categorical variable with more than two classes in Appendix~\ref{app:categorical}, and assess the sensitivity of VS accuracy to model size in Appendix~\ref{app:model_size}.

\subsection{Scaling to Longer Reasoning Chains}

\textsc{ProntoQA} structures each reasoning trace into discrete hops: one hop corresponds to the application of a deduction rule (modus ponens), which itself is broken down in more than one statement. For example, a 5-hop can involve up to 13 reasoning steps. The number of hops allows us to control the length of the reasoning chain and therefore the complexity of inference.

\begin{figure}[t]
    \centering
    \includegraphics[width=1.0\linewidth]{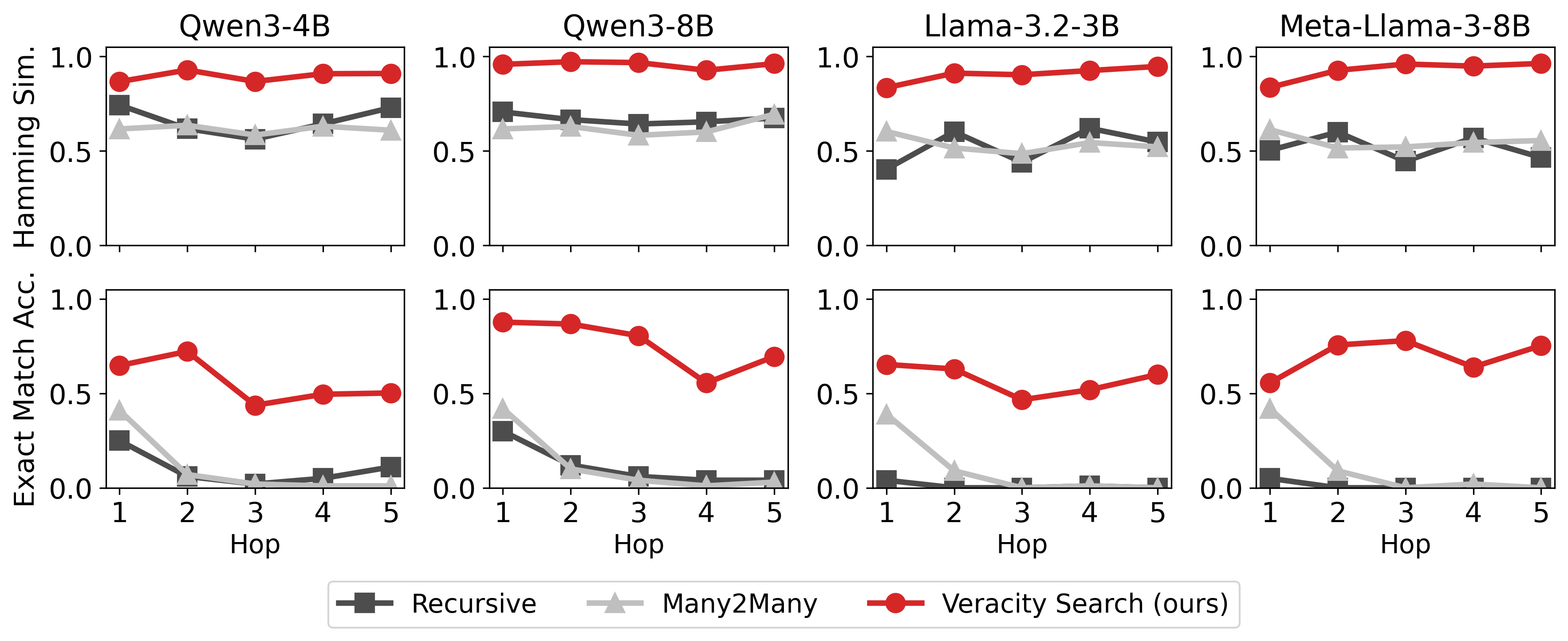}
    \caption{Veracity inference evaluation for different number of hops in \textsc{ProntoQA}. \textbf{Top row}: Mean Hamming Similarity; \textbf{Bottom row}: Mean Exact Match Accuracy. Mean is estimated using 100 test samples. }
    \label{fig:hops}
\end{figure}

We evaluated four LMs (Qwen3-4B, Qwen3-8B, Llama-3.2-3B, Llama-3-8B) on logical reasoning problems ranging from 1 to 5 hops, uniformly flipping half of the ground-truth statements in the ground-truth CoT $z^*$ to produce corrupted versions $\tilde{z}$. Fig.~\ref{fig:hops} shows how well each method recovers the ground-truth veracity vector. In addition to Hamming Similarity, we also measured Exact Match Accuracy, which is equal to 1 if the predicted $v_z$ matches the ground truth $v^*_z$, and 0 otherwise.

All methods maintained nearly constant Hamming similarity throughout the range of hops, with VS consistently above 0.85, outperforming baselines by 20–25 points. Exact-match accuracy inevitably decayed with increased hops, as the probability of correctly predicting all errors shrinks exponentially in $|z|$. Nevertheless, VS maintained relatively stable Exact-match accuracy even in 5-hop scenarios where the baselines already demonstrate a significant performance gap, failing to identify any error.

\subsection{Ablation Study for Search Hyperparameters}
\label{sec:ablation}

\begin{figure}[t]
  \centering
  \begin{subfigure}[b]{0.32\textwidth}
    \centering
    \includegraphics[width=\linewidth]{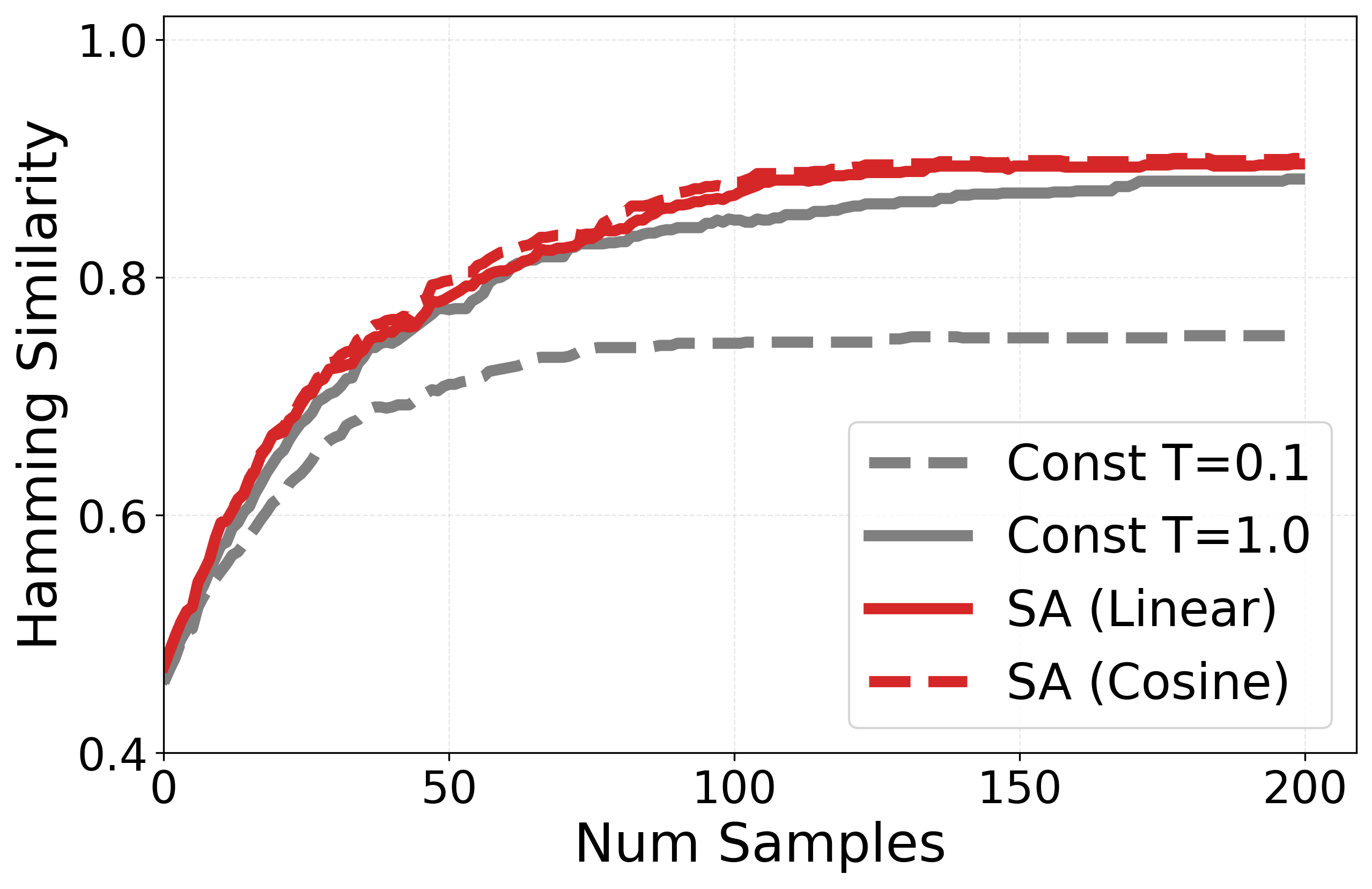}
    \caption{Simulated Annealing (SA)} \label{fig:abl1}
  \end{subfigure}\hfill
  \begin{subfigure}[b]{0.32\textwidth}
    \centering
    \includegraphics[width=\linewidth]{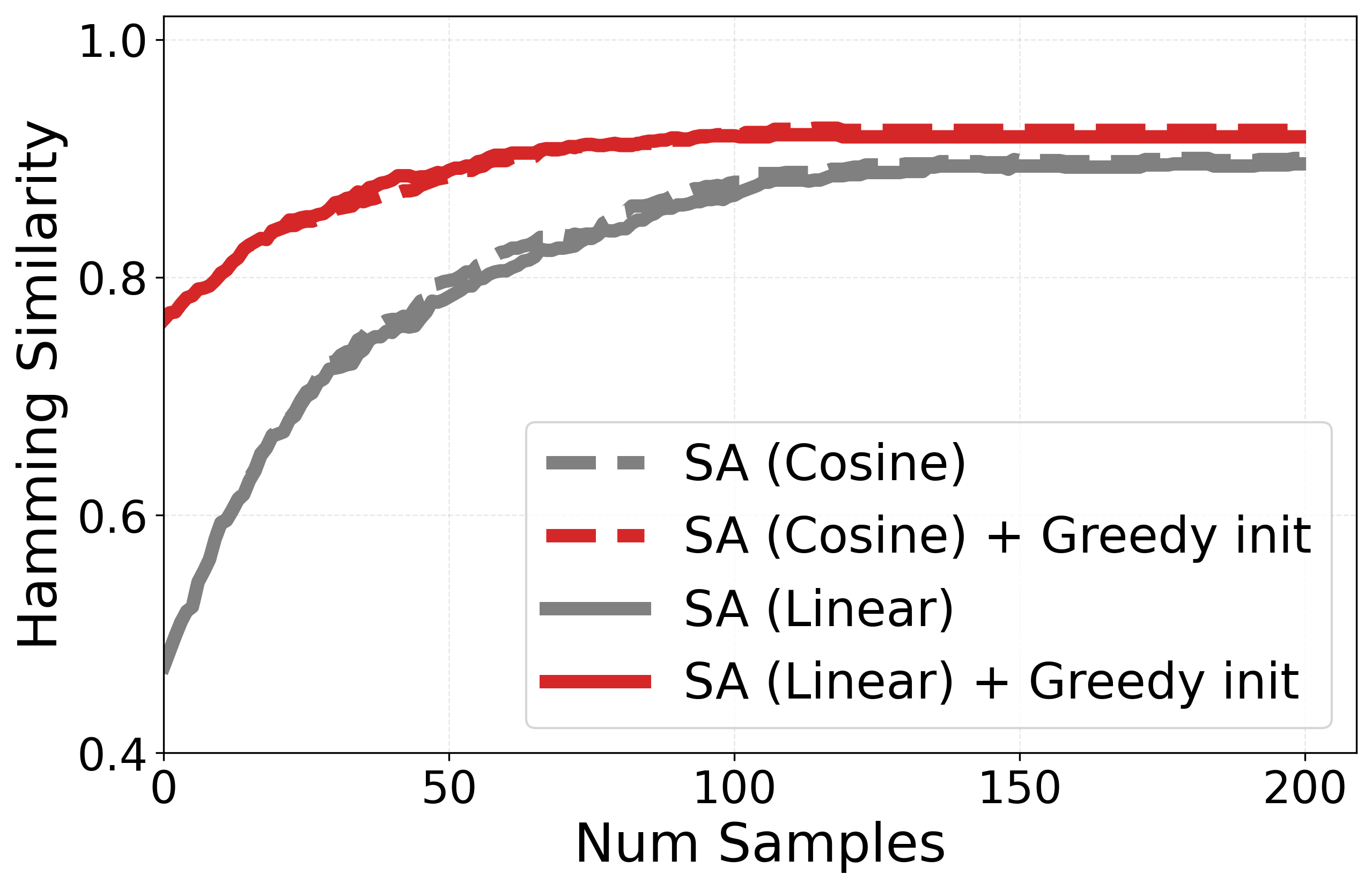}
    \caption{Greedy-tree init.} \label{fig:abl2}
  \end{subfigure}\hfill
  \begin{subfigure}[b]{0.32\textwidth}
    \centering
    \includegraphics[width=\linewidth]{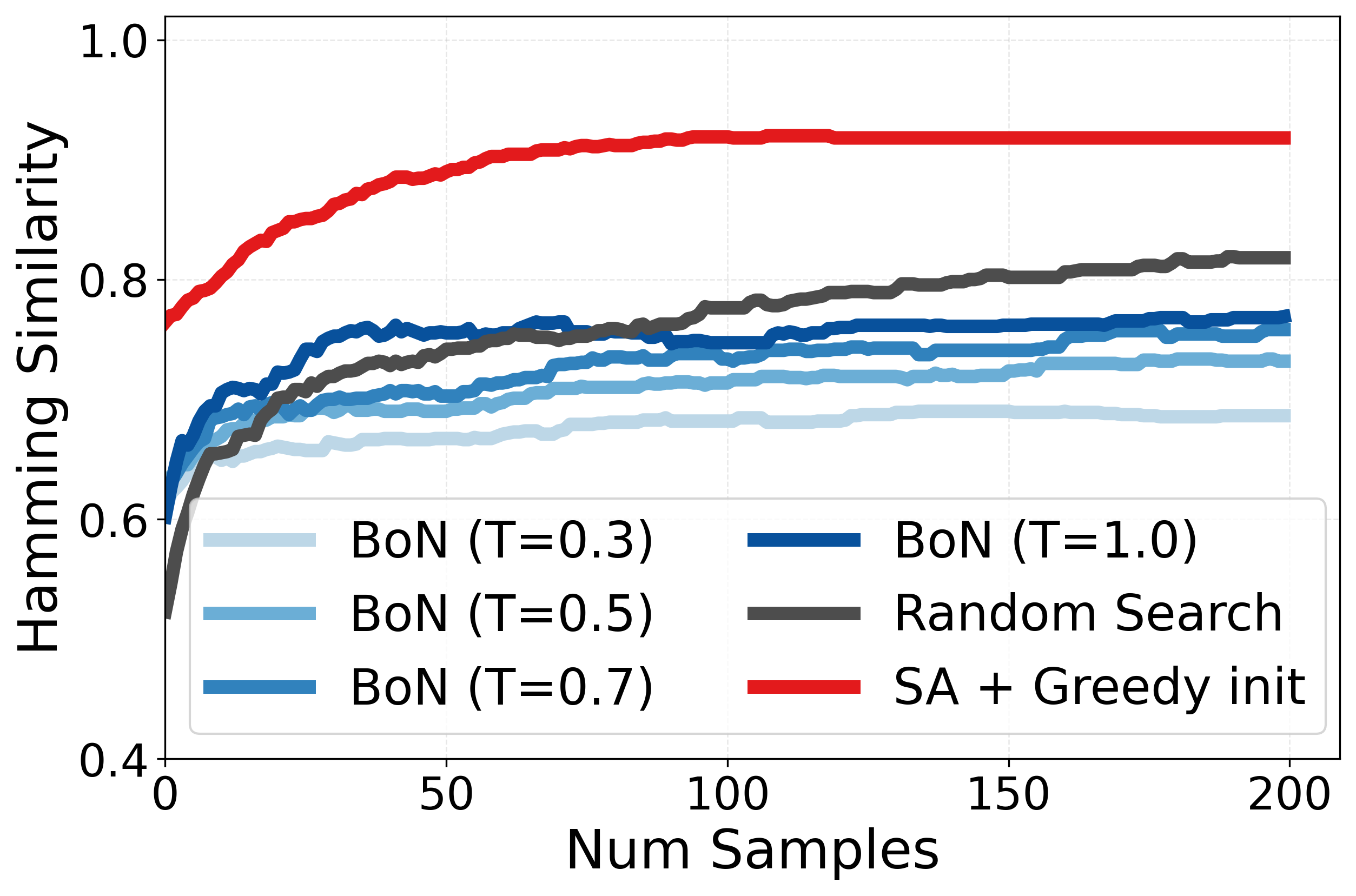}
    \caption{SA vs other search algorithms} \label{fig:abl3}
  \end{subfigure}
  \caption{Ablation study for search hyperparameters. SA: Simulated Annealing; Const: Constant Temperature; BoN($T$): Best-of-$N$ baseline using LM proposals sampled at temperature $T$. Mean value is estimated over 100 test samples from \textsc{ProntoQA}.}
  \label{fig:ablation_study}
\end{figure}

We conduct three ablations to analyze key design choices in VS, averaging results over identical \textsc{ProntoQA} splits (see Figure~\ref{fig:ablation_study}): (1) \textbf{Simulated annealing} (Figure~\ref{fig:abl1}): Linear and cosine annealing ($T_1 = 1/\beta_!=2$, $T_M = 1/\beta_M=0.1$) yield similar results and slightly outperform constant temperature ($T=1$), demonstrating the advantage of gradually increasing the inverse temperature to escape local optima for improved approximate global maximization of the proxy-reward. A constant inverse temperature ($T=0.1$) results in getting stuck in local optima; (2) \textbf{Greedy-tree initialization} (Figure~\ref{fig:abl2}): The greedy initialization method using simple tree search (\S~\ref{sec:greedy-search}) significantly boosts sample efficiency by starting from a high-quality initial solution; (3) \textbf{Comparison with other search algorithms} (Figure~\ref{fig:abl3}): Our Metropolis algorithm with Simulated Annealing clearly outperforms uniform random bit-flips (random search) and Best-of-$N$ (BoN) LM-generated proposals. Random search wastes resources, while BoN lacks local exploration, underscoring the necessity of structure-aware local moves.

Collectively, these ablations confirm that simulated annealing, principled initialization, and structured local moves significantly enhance the performance of VS. In Appendix~\ref{app:block}, we extend our method to block Metropolis, allowing for multi-bit flips in settings where more complex joint dependencies exist between veracity variables.

\subsection{Evaluation of AVI}

\label{sec:exp_amortized}

Qwen3-4B and Qwen3-8B models were fine-tuned using pseudo-labels generated by VS, for 5,000 contexts ($(x,y^*,z)$ tuples) in a 4-hop \textsc{ProntoQA} training dataset, as described in Appendix~\ref{app:implementation}. 

\paragraph{Veracity inference.} Table~\ref{tab:your_label1} reports the Hamming similarity between $v_z$ selected by the AVI and the
ground-truth veracity vector $v^{*}_{z}$ in a test set of 100 \textsc{ProntoQA} examples.
Despite being fine-tuned only on $4$-hop proofs, the model generalizes
to unseen chain lengths.
Across both Qwen backbones, Hamming similarity increases by
$\approx\!\!15$–$25$ points relative to the strongest one-shot
baseline, closing most of the gap to the optimal similarity of~1.0.

\begin{table}[t]
  \centering
  \caption{
  Hamming similarity between predicted veracity from the AVI and ground truth labels $v^*_z$. 
  Mean similarity is computed over 100 test samples from 3,4, and 5-hop \textsc{ProntoQA}.  }
  \resizebox{0.95\linewidth}{!}{%
    \begin{tabular}{lccc|ccc}
      \toprule
      \multirow{2}{*}{Base LLM} 
        & \multicolumn{3}{c}{Qwen 4B} 
        & \multicolumn{3}{c}{Qwen 8B} \\
      \cmidrule(lr){2-4} \cmidrule(lr){5-7}
        & 3-hop & 4-hop & 5-hop 
        & 3-hop & 4-hop & 5-hop \\
      \midrule
     
Many2Many
        & 0.710$\pm$0.197& 0.779$\pm$0.131 & 0.684$\pm$0.131
        & 0.663$\pm$0.146 & 0.643$\pm$0.223& 0.665$\pm$0.176 \\
AVI (ours)
        & \textbf{0.886}$\pm$0.143 & \textbf{0.921}$\pm$0.010 &  \textbf{0.913}$\pm$0.108
        & \textbf{0.956}$\pm$0.008 & \textbf{0.967}$\pm$0.069& \textbf{0.955}$\pm$0.081 \\
      \bottomrule
    \end{tabular}%
  }
  \label{tab:your_label1}
\end{table}

\paragraph{Effect on downstream reasoning.}
In \textsc{ProntoQA}, we first generate correct proofs $z^*$, then inject errors by randomly negating a subset of statements to create a corrupted reasoning chain $\tilde{z}$.
When an LM conditions directly on this flawed $\tilde{z}$, its ability to predict the correct answer $y^*$ drops sharply. A simple self-correction method leveraging AVI works as follows: first, incorrect statements are identified by the AVI, then these statements are negated to form a corrected chain $z'$ (e.g.\ replacing ``\emph{every impus is temperate}'' with ``\emph{not every impus is temperate}''). This process is illustrated in~Fig. \ref{fig:figure1}.

Table~\ref{tab:synthetic_cot_correction} compares three scenarios:
\textbf{(a)} using the uncorrected $\tilde{z}$,
\textbf{(b)} self-correction using the Many2Many baseline, and
\textbf{(c)} employing AVI for self-correction.
Conditioning the LM on $z'$ boosts the conditional probability of the true answer by up to $25\%$ on Qwen-8B and by $10$--$12$ points on Qwen-4B.
The improvement is consistent across $3$-, $4$-, and $5$-hop proofs, suggesting that simply correcting mistakes in a CoT can result in improved reasoning accuracy. We extend this evaluation to LM-generated reasoning chains $z$ (with no synthetic corruption) in Appendix~\ref{app:self-refine}.

\begin{table}[t]
  \centering
  \caption{Reasoning accuracy for inferring $y^*$ from \textsc{ProntoQA} problems given synthetic noisy chains. Average accuracy and standard deviation computed over 100 problems.}
  \resizebox{0.95\linewidth}{!}{%
    \begin{tabular}{lccc|ccc}
      \toprule
      \multirow{2}{*}{Method} & \multicolumn{3}{c}{Qwen 4B} & \multicolumn{3}{c}{Qwen 8B} \\
      \cmidrule(lr){2-4}\cmidrule(lr){5-7}
      & 3-hop & 4-hop & 5-hop & 3-hop & 4-hop & 5-hop \\
      \midrule
      No Correction           & 0.60$\pm$0.05 & 0.52$\pm$0.05 & 0.59$\pm$0.05 & 0.54$\pm$0.05 & 0.65$\pm$0.05 & 0.52$\pm$0.05 \\
      Self Correction         & 0.54$\pm$0.05 & 0.60$\pm$0.05 & 0.48$\pm$0.05 & 0.54$\pm$0.05 & 0.58$\pm$0.05 & 0.46$\pm$0.05 \\
      AVI (ours) & \textbf{0.68$\pm$0.05} & \textbf{0.72$\pm$0.04} & \textbf{0.77$\pm$0.04} & \textbf{0.87$\pm$0.03} & \textbf{0.85$\pm$0.04} & \textbf{0.81$\pm$0.04} \\
      \bottomrule
    \end{tabular}%
  }
  \label{tab:synthetic_cot_correction}
\end{table}

\begin{figure}[t]
  \centering
  \begin{subfigure}[b]{0.48\textwidth}
    \centering
    \includegraphics[width=\linewidth]{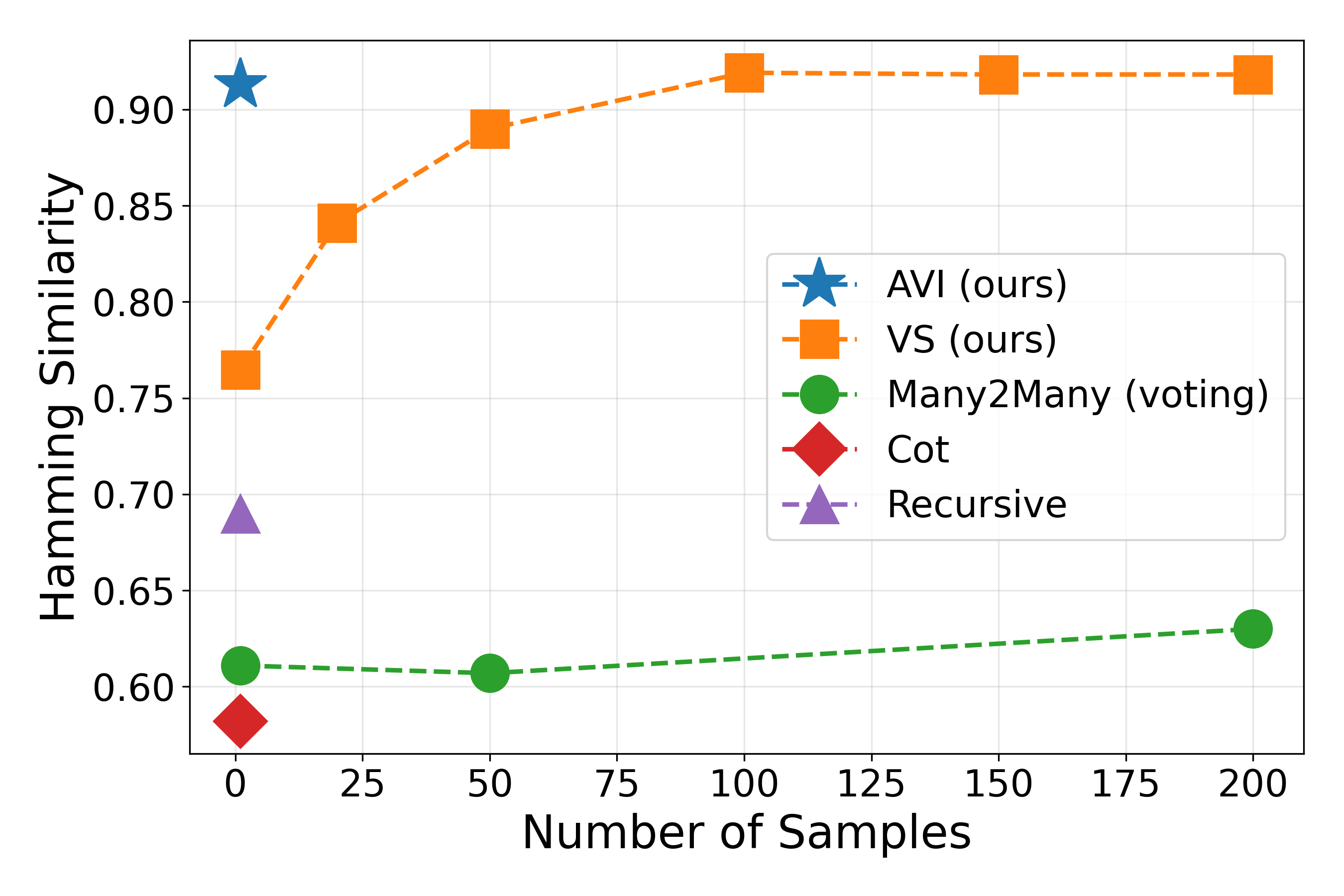}
  \end{subfigure}
  \begin{subfigure}[b]{0.48\textwidth}
    \centering
    \includegraphics[width=\linewidth]{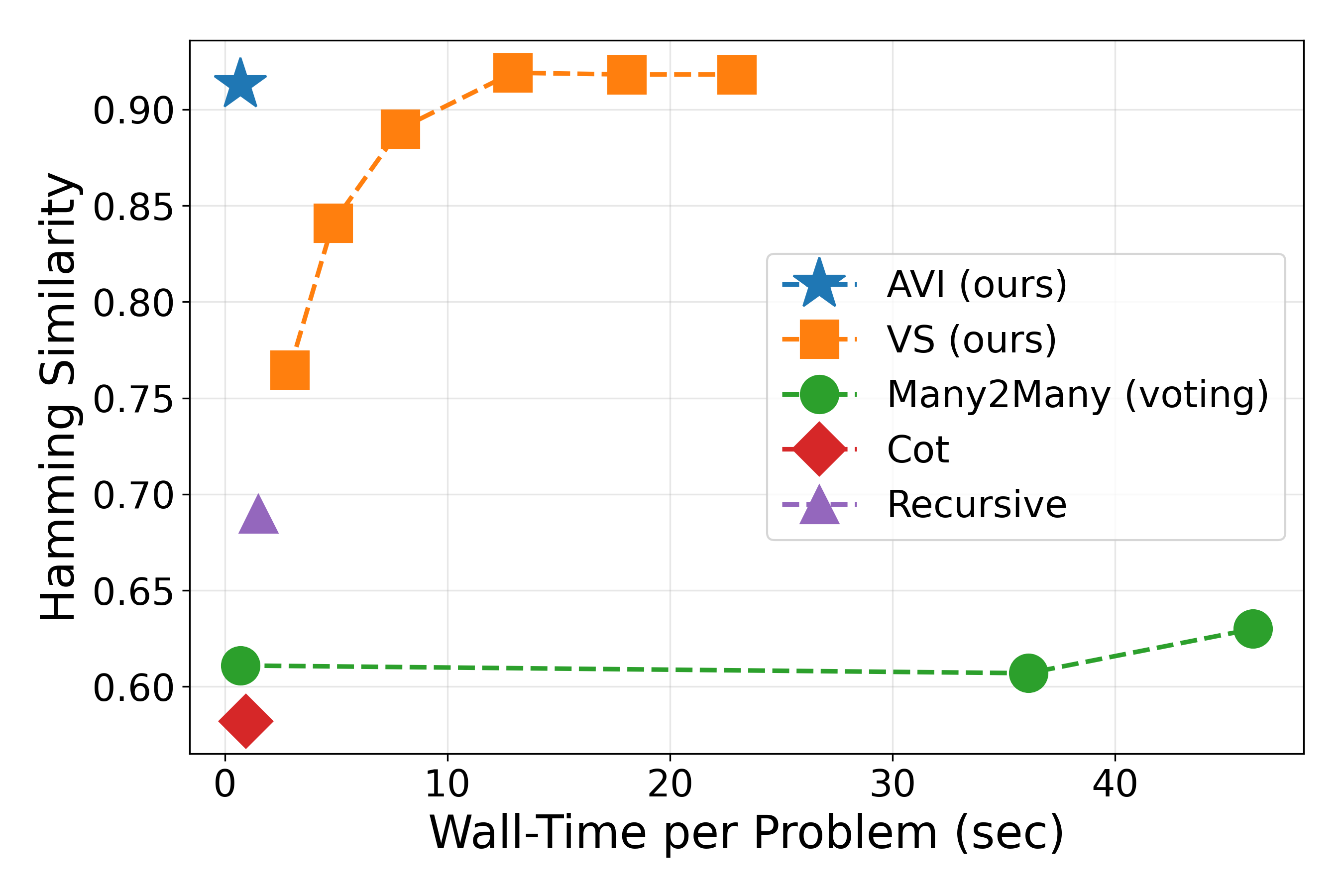}
  \end{subfigure}
  \caption{Test-time inference efficiency of Qwen3-4B on 100 problems $(x,z,y^*)$ from 5-hop \textsc{ProntoQA} for inferring $v_z$. Inference wall time is shown as the number of samples per problem scales.}
  \label{fig:inference-eff}
  \vspace{-1em}
\end{figure}

Inference-time self-improvement methods are related to self-correction, and aim to improve reasoning through resampling and editing, occasionally utilising a distinct feedback module for guiding correction. Our approach is complementary to these self-improvement methods, as it focuses on identifying statements that need correction, and would therefore fit naturally as a feedback module. In Appendix~\ref{app:self-refine}, we illustrate how AVI can be used as the feedback module in the self-improvment method Self-Refine~\citep{madaan2023self}, replacing the few-shot prompted LM (corresponding to our baseline Many2Many) that is typically used for this purpose. 
We find that AVI continues to outperform baselines for this type of application, but note that the magnitude of improvement in reasoning performance is smaller than the gain observed in terms of veracity inference accuracy in the preceding experiments. This suggests that verification alone is not the only bottleneck in self-correction/improvement frameworks~\citep{Zhang2024SmallLM}, and more work is needed to better understand how to make the best use of a stronger verification signal to guide CoT-(re)generation.

\subsection{Sample efficiency and wall-time}
\label{app:inference-eff}

Finally, we evaluate the sample-efficiency and inference-time requirements of our proposed method compared to the same baselines from previous experiments. We use 100 examples from the 5-hop \textsc{ProntoQA} dataset and run all experiments on the same NVIDIA RTX 6000 ADA (48GB) GPU.

As depicted in Figure~\ref{fig:inference-eff} (left), VS outperforms all baseline methods in terms of sample efficiency, even when limited to a single sample using greedy tree-based initialization. Performance scales favorably, reaching optimal results at around 100 samples. The zero-shot (1-sample) AVI provides Pareto-optimal performance, comparable to the 100-sample iteration of VS.
As shown in Figure~\ref{fig:inference-eff} (right), VS also uses less inference time per sample compared to other baselines. This efficiency stems from the parallelized reward computations using LM likelihoods evaluated on sampled sequences, eliminating the need for sequential token-level decoding that other baselines require. We include a complementary analysis of the inference-time cost of our approach in Appendix~\ref{app:complexity}.

\section{Conclusion}

We introduced a framework for stepwise error-identification that extends the latent variable representation of a CoT by disentangling its identity $z$ from its veracity $V_z$.
We proposed a discrete search algorithm, Veracity Search (VS), to efficiently search over boolean veracity vectors, and proposed a method (Amortized Veracity Inference; AVI) for fine-tuning an LM with pseudo-labels from VS, allowing us to apply VS in contexts where the final answer is unknown. 
Empirically, we found that VS outperforms in-context learning baselines in logical, mathematical, and commonsense reasoning tasks when assessing the accuracy of the predicted veracity of individual reasoning steps, and that AVI can complement downstream tasks that utilize veracity labels as feedback for correction or for guiding the re-generation of erroneous reasoning chains towards ones that increase reasoning performance. 

There are important limitations to our work. 
The performance of VS can be sensitive to the distribution of errors, and most of our analysis pertains to artificially corrupted CoTs. Preliminary results suggest that these techniques also work for naturally occurring errors (\S\ref{app:different_patterns}), but these claims should be tested in a broader range of dedicated experiments. 
Ultimately, important applications of error-identification (such as self-correction) often require more than flagging mistakes: inferring veracity $v_z$ of a CoT does not render the identity $z$ of the reasoning steps unimportant. In the appendix (\S\ref{app:self-refine}) we prototype a way of integrating AVI with self-improvement methods to resample erroneous statements in $z$, but designing more powerful reasoning systems that combine veracity inference with iterative test-time correction schemes remains an important avenue for future work. Our method could also provide a new training signal for reasoning models, for example by using EM to jointly update the generative model with veracity samples obtained with AVI, or by providing a veracity-specific process-level training signal for PRMs in label-scarcity scenarios.

\clearpage

\section*{Acknowledgments}
We thank researchers at Mila (Joumana Ghosn, Pierre-Luc St-Charles, Marc-Antoine Rondeau, Mohsin Hasan, Siddarth Venkatraman, Yaroslav Kivva) and KAIST (Junyeop Baek, Doojin Baek) for insightful discussions and assistance with this project. This research is supported by the Canadian AI Safety Institute Research Program at CIFAR through a Catalyst award. The research was enabled in part by computational resources provided by the Digital Research
Alliance of Canada (\url{https://alliancecan.ca}), Mila (\url{https://mila.quebec}), and NVIDIA. The authors acknowledge funding from CIFAR, NSERC and the Future of Life Institute. Minsu Kim acknowledges funding from KAIST Jang Young Sil Fellow Program. Jean-Pierre Falet is supported by a Doctoral Vanier Canada Graduate Scholarship (FRN: CGV-192746). Moksh Jain acknowledges funding from a FRQNT Doctoral Fellowship (\url{https://doi.org/10.69777/366694}). Minsu Kim, Sungsoo Ahn, and Sungjin Ahn were supported by GRDC Cooperative Hub Program (RS-2024-00436165) through the National Research Foundation of Korea (NRF).

\bibliography{iclr2026_conference}

@inproceedings{Zhang2024SmallLM,
  title={Small Language Models Need Strong Verifiers to Self-Correct Reasoning},
  author={Yunxiang Zhang and Muhammad Khalifa and Lajanugen Logeswaran and Jaekyeom Kim and Moontae Lee and Honglak Lee and Lu Wang},
  booktitle={Annual Meeting of the Association for Computational Linguistics},
  year={2024},
  url={https://api.semanticscholar.org/CorpusID:269430362}
}

@article{yang-etal-2022-generating,
  title={Generating natural language proofs with verifier-guided search},
  author={Yang, Kaiyu and Deng, Jia and Chen, Danqi},
  journal={Conference on Empirical Methods in Natural Language Processing (EMNLP)},
  year={2022}
}

@article{zhou2024can,
  title={Can language models perform robust reasoning in chain-of-thought prompting with noisy rationales?},
  author={Zhou, Zhanke and Tao, Rong and Zhu, Jianing and Luo, Yiwen and Wang, Zengmao and Han, Bo},
  journal={Neural Information Processing Systems (NeurIPS)},
  year={2024}
}

@article{bengio2021flow,
    title={Flow Network based Generative Models for Non-Iterative Diverse Candidate Generation},
    author={Emmanuel Bengio and Moksh Jain and Maksym Korablyov and Doina Precup and Yoshua Bengio},
    journal={Neural Information Processing Systems (NeurIPS)},
    year={2021}
}

@article{hu2023amortizing,
    title={Amortizing intractable inference in large language models},
    author={Hu, Edward J. and Jain, Moksh and Elmoznino, Eric and  Kaddar, Younesse and Lajoie, Guillaume and Bengio, Yoshua and Malkin, Nikolay},
    journal={International Conference on Learning Representations (ICLR)},
    year={2024}
}

@inproceedings{rajani-etal-2019-explain,
    title = "Explain Yourself! Leveraging Language Models for Commonsense Reasoning",
    author = "Rajani, Nazneen Fatema  and
      McCann, Bryan  and
      Xiong, Caiming  and
      Socher, Richard",
    editor = "Korhonen, Anna  and
      Traum, David  and
      M{\`a}rquez, Llu{\'i}s",
    booktitle = "Proceedings of the 57th Annual Meeting of the Association for Computational Linguistics",
    month = jul,
    year = "2019",
    address = "Florence, Italy",
    publisher = "Association for Computational Linguistics",
    url = "https://aclanthology.org/P19-1487/",
    doi = "10.18653/v1/P19-1487",
    pages = "4932--4942",
}

@article{
lightman2024lets,
title={Let's Verify Step by Step},
author={Hunter Lightman and Vineet Kosaraju and Yuri Burda and Harrison Edwards and Bowen Baker and Teddy Lee and Jan Leike and John Schulman and Ilya Sutskever and Karl Cobbe},
journal={International Conference on Learning Representations (ICLR)},
year={2024}}

@article{cobbe2021training,
  title={Training verifiers to solve math word problems},
  author={Cobbe, Karl and Kosaraju, Vineet and Bavarian, Mohammad and Chen, Mark and Jun, Heewoo and Kaiser, Lukasz and Plappert, Matthias and Tworek, Jerry and Hilton, Jacob and Nakano, Reiichiro and others},
  journal={arXiv preprint arXiv:2110.14168},
  year={2021}
}

@article{shinn2023reflexion,
  title={Reflexion: Language agents with verbal reinforcement learning},
  author={Shinn, Noah and Cassano, Federico and Gopinath, Ashwin and Narasimhan, Karthik and Yao, Shunyu},
  journal={Neural Information Processing Systems (NeurIPS)},
  year={2023}
}

@article{weng2022large,
  title={Large language models are better reasoners with self-verification},
  author={Weng, Yixuan and Zhu, Minjun and Xia, Fei and Li, Bin and He, Shizhu and Liu, Shengping and Sun, Bin and Liu, Kang and Zhao, Jun},
  journal={Findings of the Association for Computational Linguistics: EMNLP 2023},
  year={2023}
}

@article{li2023self,
  title={Self-checker: Plug-and-play modules for fact-checking with large language models},
  author={Li, Miaoran and Peng, Baolin and Galley, Michel and Gao, Jianfeng and Zhang, Zhu},
  journal={Findings of the Association for Computational Linguistics: NAACL 2024},
  year={2023}
}

@article{wei2022chain,
  title={Chain-of-thought prompting elicits reasoning in large language models},
  author={Wei, Jason and Wang, Xuezhi and Schuurmans, Dale and Bosma, Maarten and Xia, Fei and Chi, Ed and Le, Quoc V and Zhou, Denny and others},
  journal={Neural Information Processing Systems (NeurIPS)},
  year={2022}
}

@article{nye2021show,
  title={Show your work: Scratchpads for intermediate computation with language models},
  author={Nye, Maxwell and Andreassen, Anders Johan and Gur-Ari, Guy and Michalewski, Henryk and Austin, Jacob and Bieber, David and Dohan, David and Lewkowycz, Aitor and Bosma, Maarten and Luan, David and others},
 journal={arXiv preprint arXiv:2112.00114},
  year={2021}
}

@article{kojima2022large,
  title={Large language models are zero-shot reasoners},
  author={Kojima, Takeshi and Gu, Shixiang Shane and Reid, Machel and Matsuo, Yutaka and Iwasawa, Yusuke},
  journal={Neural Information Processing Systems (NeurIPS)},
  year={2022}
}

@article{turpin2023language,
  title={Language models don't always say what they think: Unfaithful explanations in chain-of-thought prompting},
  author={Turpin, Miles and Michael, Julian and Perez, Ethan and Bowman, Samuel},
  journal={Neural Information Processing Systems (NeurIPS)},
  year={2023}
}

@article{perez2023discovering,
  title={Discovering language model behaviors with model-written evaluations},
  author={Perez, Ethan and Ringer, Sam and Lukosiute, Kamile and Nguyen, Karina and Chen, Edwin and Heiner, Scott and Pettit, Craig and Olsson, Catherine and Kundu, Sandipan and Kadavath, Saurav and others},
  journal={Findings of the Association for Computational Linguistics: ACL 2023},
  year={2023}
}

@article{zhang2023language,
  title={How language model hallucinations can snowball},
  author={Zhang, Muru and Press, Ofir and Merrill, William and Liu, Alisa and Smith, Noah A},
  journal={International Conference on Machine Learning (ICML)},
  year={2024}
}

@article{camburu2018esnli,
  title={e-snli: Natural language inference with natural language explanations},
  author={Camburu, Oana-Maria and Rockt{\"a}schel, Tim and Lukasiewicz, Thomas and Blunsom, Phil},
  journal={Neural Information Processing Systems (NeurIPS)},
  year={2018}
}

@article{golovneva2022roscoe,
  title={Roscoe: A suite of metrics for scoring step-by-step reasoning},
  author={Golovneva, Olga and Chen, Moya and Poff, Spencer and Corredor, Martin and Zettlemoyer, Luke and Fazel-Zarandi, Maryam and Celikyilmaz, Asli},
  journal={International Conference on Learning Representations (ICLR)},
  year={2023}
}

@article{zelikman2022star,
  title={STaR: Bootstrapping reasoning with reasoning},
  author={Zelikman, Eric and Wu, Yuhuai and Mu, Jesse and Goodman, Noah},
  journal={Neural Information Processing Systems (NeurIPS)},
  year={2022}
}

@article{ji2023survey,
  title={Survey of hallucination in natural language generation},
  author={Ji, Ziwei and Lee, Nayeon and Frieske, Rita and Yu, Tiezheng and Su, Dan and Xu, Yan and Ishii, Etsuko and Bang, Ye Jin and Madotto, Andrea and Fung, Pascale},
  journal={ACM computing surveys},
  volume={55},
  number={12},
  pages={1--38},
  year={2023},
  publisher={ACM New York, NY}
}

@article{bang2023multitask,
  title={A multitask, multilingual, multimodal evaluation of chatgpt on reasoning, hallucination, and interactivity},
  author={Bang, Yejin and Cahyawijaya, Samuel and Lee, Nayeon and Dai, Wenliang and Su, Dan and Wilie, Bryan and Lovenia, Holy and Ji, Ziwei and Yu, Tiezheng and Chung, Willy and others},
  journal={arXiv preprint arXiv:2302.04023},
  year={2023}
}

@article{yao2023tree,
  title={Tree of thoughts: Deliberate problem solving with large language models},
  author={Yao, Shunyu and Yu, Dian and Zhao, Jeffrey and Shafran, Izhak and Griffiths, Tom and Cao, Yuan and Narasimhan, Karthik},
  journal={Neural Information Processing Systems (NeurIPS)},
  year={2023}
}

@article{wang2023selfconsistency,
title={Self-Consistency Improves Chain of Thought Reasoning in Language Models},
author={Xuezhi Wang and Jason Wei and Dale Schuurmans and Quoc V Le and Ed H. Chi and Sharan Narang and Aakanksha Chowdhery and Denny Zhou},
journal={International Conference on Learning Representations (ICLR)},
year={2023},
}

@article{chern2023factool,
  title={FacTool: Factuality Detection in Generative AI--A Tool Augmented Framework for Multi-Task and Multi-Domain Scenarios},
  author={Chern, I and Chern, Steffi and Chen, Shiqi and Yuan, Weizhe and Feng, Kehua and Zhou, Chunting and He, Junxian and Neubig, Graham and Liu, Pengfei and others},
  journal={arXiv preprint arXiv:2307.13528},
  year={2023}
}

@article{min2023factscore,
  title={Factscore: Fine-grained atomic evaluation of factual precision in long form text generation},
  author={Min, Sewon and Krishna, Kalpesh and Lyu, Xinxi and Lewis, Mike and Yih, Wen-tau and Koh, Pang Wei and Iyyer, Mohit and Zettlemoyer, Luke and Hajishirzi, Hannaneh},
  journal={arXiv preprint arXiv:2305.14251},
  year={2023}
}

@article{madaan2023self,
  title={Self-refine: Iterative refinement with self-feedback},
  author={Madaan, Aman and Tandon, Niket and Gupta, Prakhar and Hallinan, Skyler and Gao, Luyu and Wiegreffe, Sarah and Alon, Uri and Dziri, Nouha and Prabhumoye, Shrimai and Yang, Yiming and others},
  journal={Neural Information Processing Systems (NeurIPS)},
  year={2023}
}

@article{huang2022large,
  title={Large language models can self-improve},
  author={Huang, Jiaxin and Gu, Shixiang Shane and Hou, Le and Wu, Yuexin and Wang, Xuezhi and Yu, Hongkun and Han, Jiawei},
  journal={Conference on Empirical Methods in Natural Language Processing (EMNLP)},
  year={2023}
}

@article{pan2023automatically,
    title = "Automatically Correcting Large Language Models: Surveying the Landscape of Diverse Automated Correction Strategies",
    author = "Pan, Liangming  and
      Saxon, Michael  and
      Xu, Wenda  and
      Nathani, Deepak  and
      Wang, Xinyi  and
      Wang, William Yang",
    journal = "Transactions of the Association for Computational Linguistics",
    volume = "12",
    year = "2024",
    address = "Cambridge, MA",
    publisher = "MIT Press",
    url = "https://aclanthology.org/2024.tacl-1.27/",
    doi = "10.1162/tacl_a_00660",
    pages = "484--506"
}

@article{cui2025process,
  title={Process reinforcement through implicit rewards},
  author={Cui, Ganqu and Yuan, Lifan and Wang, Zefan and Wang, Hanbin and Zhang, Yuchen and Chen, Jiacheng and Li, Wendi and He, Bingxiang and Fan, Yuchen and Yu, Tianyu and others},
  journal={arXiv preprint arXiv:2502.01456},
  year={2025}
}

@article{zha2025rl,
  title={RL Tango: Reinforcing Generator and Verifier Together for Language Reasoning},
  author={Zha, Kaiwen and Gao, Zhengqi and Shen, Maohao and Hong, Zhang-Wei and Boning, Duane S and Katabi, Dina},
  journal={arXiv preprint arXiv:2505.15034},
  year={2025}
}

@inproceedings{
yuan2025free,
title={Free Process Rewards without Process Labels},
author={Lifan Yuan and Wendi Li and Huayu Chen and Ganqu Cui and Ning Ding and Kaiyan Zhang and Bowen Zhou and Zhiyuan Liu and Hao Peng},
booktitle={Forty-second International Conference on Machine Learning},
year={2025},
url={https://openreview.net/forum?id=8ThnPFhGm8}
}

@article{jacovi2024chain,
  title={A chain-of-thought is as strong as its weakest link: A benchmark for verifiers of reasoning chains},
  author={Jacovi, Alon and Bitton, Yonatan and Bohnet, Bernd and Herzig, Jonathan and Honovich, Or and Tseng, Michael and Collins, Michael and Aharoni, Roee and Geva, Mor},
  journal={Annual Meeting of the Association for Computational Linguistics (ACL)},
  year={2024}
}

@article{manakul2023selfcheckgpt,
  title={Selfcheckgpt: Zero-resource black-box hallucination detection for generative large language models},
  author={Manakul, Potsawee and Liusie, Adian and Gales, Mark JF},
  journal={Conference on Empirical Methods in Natural Language Processing (EMNLP)},
  year={2023}
}

@article{xie2023self,
  title={Self-evaluation guided beam search for reasoning},
  author={Xie, Yuxi and Kawaguchi, Kenji and Zhao, Yiran and Zhao, James Xu and Kan, Min-Yen and He, Junxian and Xie, Michael},
  journal={Neural Information Processing Systems (NeurIPS)},
  year={2023}
}

@article{saparov2022language,
  title={Language models are greedy reasoners: A systematic formal analysis of chain-of-thought},
  author={Saparov, Abulhair and He, He},
  journal={International Conference on Learning Representations (ICLR)},
  year={2022}
}

@article{guo2025deepseek,
  title={Deepseek-r1: Incentivizing reasoning capability in llms via reinforcement learning},
  author={Guo, Daya and Yang, Dejian and Zhang, Haowei and Song, Junxiao and Zhang, Ruoyu and Xu, Runxin and Zhu, Qihao and Ma, Shirong and Wang, Peiyi and Bi, Xiao and others},
  journal={arXiv preprint arXiv:2501.12948},
  year={2025}
}

@article{phan2023training,
  title={Training chain-of-thought via latent-variable inference},
  author={Phan, Du and Hoffman, Matthew Douglas and Dohan, David and Douglas, Sholto and Le, Tuan Anh and Parisi, Aaron and Sountsov, Pavel and Sutton, Charles and Vikram, Sharad and A Saurous, Rif},
  journal={Neural Information Processing Systems (NeurIPS)},
  year={2023}
}

@article{uesato2022solving,
  title={Solving math word problems with process-and outcome-based feedback},
  author={Uesato, Jonathan and Kushman, Nate and Kumar, Ramana and Song, Francis and Siegel, Noah and Wang, Lisa and Creswell, Antonia and Irving, Geoffrey and Higgins, Irina},
  journal={arXiv preprint arXiv:2211.14275},
  year={2022}
}

@article{gulcehre2023reinforced,
  title={Reinforced self-training (rest) for language modeling},
  author={Gulcehre, Caglar and Paine, Tom Le and Srinivasan, Srivatsan and Konyushkova, Ksenia and Weerts, Lotte and Sharma, Abhishek and Siddhant, Aditya and Ahern, Alex and Wang, Miaosen and Gu, Chenjie and others},
  journal={arXiv preprint arXiv:2308.08998},
  year={2023}
}

@inproceedings{xue2023dynamic,
  title={Dynamic voting for efficient reasoning in large language models},
  author={Xue, Mingfeng and Liu, Dayiheng and Lei, Wenqiang and Ren, Xingzhang and Yang, Baosong and Xie, Jun and Zhang, Yidan and Peng, Dezhong and Lv, Jiancheng},
  booktitle={Findings of the Association for Computational Linguistics: EMNLP 2023},
  pages={3085--3104},
  year={2023}
}

@article{xu2025softcot,
  title={Softcot: Soft chain-of-thought for efficient reasoning with llms},
  author={Xu, Yige and Guo, Xu and Zeng, Zhiwei and Miao, Chunyan},
  journal={arXiv preprint arXiv:2502.12134},
  year={2025}
}

@article{yu2024flow,
  title={Flow of Reasoning: Training LLMs for Divergent Problem Solving with Minimal Examples},
  author={Yu, Fangxu and Jiang, Lai and Kang, Haoqiang and Hao, Shibo and Qin, Lianhui},
  journal={arXiv preprint arXiv:2406.05673},
  year={2024}
}

@article{luo2024improve,
  title={Improve mathematical reasoning in language models by automated process supervision},
  author={Luo, Liangchen and Liu, Yinxiao and Liu, Rosanne and Phatale, Samrat and Guo, Meiqi and Lara, Harsh and Li, Yunxuan and Shu, Lei and Zhu, Yun and Meng, Lei and others},
  journal={arXiv preprint arXiv:2406.06592},
  year={2024}
}

@article{zhang2024accessing,
  title={Accessing gpt-4 level mathematical olympiad solutions via monte carlo tree self-refine with llama-3 8b},
  author={Zhang, Di and Huang, Xiaoshui and Zhou, Dongzhan and Li, Yuqiang and Ouyang, Wanli},
  journal={arXiv preprint arXiv:2406.07394},
  year={2024}
}

@article{xie2024monte,
  title={Monte carlo tree search boosts reasoning via iterative preference learning},
  author={Xie, Yuxi and Goyal, Anirudh and Zheng, Wenyue and Kan, Min-Yen and Lillicrap, Timothy P and Kawaguchi, Kenji and Shieh, Michael},
  journal={Neural Information Processing Systems (NeurIPS)},
  year={2024}
}

@article{bai2023qwen,
  title={Qwen technical report},
  author={Bai, Jinze and Bai, Shuai and Chu, Yunfei and Cui, Zeyu and Dang, Kai and Deng, Xiaodong and Fan, Yang and Ge, Wenbin and Han, Yu and Huang, Fei and others},
  journal={arXiv preprint arXiv:2309.16609},
  year={2023}
}

@article{qwen2.5,
    title   = {Qwen2.5 Technical Report}, 
    author  = {An Yang and Baosong Yang and Beichen Zhang and Binyuan Hui and Bo Zheng and Bowen Yu and Chengyuan Li and Dayiheng Liu and Fei Huang and Haoran Wei and Huan Lin and Jian Yang and Jianhong Tu and Jianwei Zhang and Jianxin Yang and Jiaxi Yang and Jingren Zhou and Junyang Lin and Kai Dang and Keming Lu and Keqin Bao and Kexin Yang and Le Yu and Mei Li and Mingfeng Xue and Pei Zhang and Qin Zhu and Rui Men and Runji Lin and Tianhao Li and Tingyu Xia and Xingzhang Ren and Xuancheng Ren and Yang Fan and Yang Su and Yichang Zhang and Yu Wan and Yuqiong Liu and Zeyu Cui and Zhenru Zhang and Zihan Qiu},
    journal = {arXiv preprint arXiv:2412.15115},
    year    = {2024}
}

@article{qwen2,
    title   = {Qwen2 Technical Report}, 
    author  = {An Yang and Baosong Yang and Binyuan Hui and Bo Zheng and Bowen Yu and Chang Zhou and Chengpeng Li and Chengyuan Li and Dayiheng Liu and Fei Huang and Guanting Dong and Haoran Wei and Huan Lin and Jialong Tang and Jialin Wang and Jian Yang and Jianhong Tu and Jianwei Zhang and Jianxin Ma and Jin Xu and Jingren Zhou and Jinze Bai and Jinzheng He and Junyang Lin and Kai Dang and Keming Lu and Keqin Chen and Kexin Yang and Mei Li and Mingfeng Xue and Na Ni and Pei Zhang and Peng Wang and Ru Peng and Rui Men and Ruize Gao and Runji Lin and Shijie Wang and Shuai Bai and Sinan Tan and Tianhang Zhu and Tianhao Li and Tianyu Liu and Wenbin Ge and Xiaodong Deng and Xiaohuan Zhou and Xingzhang Ren and Xinyu Zhang and Xipin Wei and Xuancheng Ren and Yang Fan and Yang Yao and Yichang Zhang and Yu Wan and Yunfei Chu and Yuqiong Liu and Zeyu Cui and Zhenru Zhang and Zhihao Fan},
    journal = {arXiv preprint arXiv:2407.10671},
    year    = {2024}
}

@article{touvron2023llama,
  title={Llama: Open and efficient foundation language models},
  author={Touvron, Hugo and Lavril, Thibaut and Izacard, Gautier and Martinet, Xavier and Lachaux, Marie-Anne and Lacroix, Timoth{\'e}e and Rozi{\`e}re, Baptiste and Goyal, Naman and Hambro, Eric and Azhar, Faisal and others},
  journal={arXiv preprint arXiv:2302.13971},
  year={2023}
}

@article{touvron2023llama2,
  title={Llama 2: Open foundation and fine-tuned chat models},
  author={Touvron, Hugo and Martin, Louis and Stone, Kevin and Albert, Peter and Almahairi, Amjad and Babaei, Yasmine and Bashlykov, Nikolay and Batra, Soumya and Bhargava, Prajjwal and Bhosale, Shruti and others},
  journal={arXiv preprint arXiv:2307.09288},
  year={2023}
}

@article{grattafiori2024llama,
  title={The llama 3 herd of models},
  author={Grattafiori, Aaron and Dubey, Abhimanyu and Jauhri, Abhinav and Pandey, Abhinav and Kadian, Abhishek and Al-Dahle, Ahmad and Letman, Aiesha and Mathur, Akhil and Schelten, Alan and Vaughan, Alex and others},
  journal={arXiv preprint arXiv:2407.21783},
  year={2024}
}

@article{snell2024scaling,
  title={Scaling llm test-time compute optimally can be more effective than scaling model parameters},
  author={Snell, Charlie and Lee, Jaehoon and Xu, Kelvin and Kumar, Aviral},
  journal={International Conference on Learning Representations (ICLR)},
  year={2025}
}

@article{openai2024o1,
  title={{OpenAI} o1 system card},
  author={OpenAI},
  journal={arXiv preprint arXiv:2412.16720},
  year={2024}
}

@article{loshchilov2017decoupled,
  title={Decoupled weight decay regularization},
  author={Loshchilov, Ilya and Hutter, Frank},
  journal={International Conference on Learning Representations (ICLR)},
  year={2019}
}

@article{pmlr-v235-zhao24c,
  title = 	 {Probabilistic Inference in Language Models via Twisted Sequential {M}onte {C}arlo},
  author =       {Zhao, Stephen and Brekelmans, Rob and Makhzani, Alireza and Grosse, Roger Baker},
  journal = 	 {International Conference on Machine Learning (ICML)},
  year = 	 {2024},
}

@article{talmor2018commonsenseqa,
  title={Commonsenseqa: A question answering challenge targeting commonsense knowledge},
  author={Talmor, Alon and Herzig, Jonathan and Lourie, Nicholas and Berant, Jonathan},
  journal={arXiv preprint arXiv:1811.00937},
  year={2018}
}
\bibliographystyle{iclr2026_conference}

\clearpage
\appendix
\section{Detailed Implementations and Experiments on Baselines}
\label{app:implementation}

\subsection{Datasets}

\paragraph{\textsc{ProntoQA}.}
We generate \( (x,y,z)\) triples with the official \textsc{ProntoQA} synthesiser, which samples first–order–logic templates with a fictional ontology.
For every instance we uniformly draw a reasoning depth \(d\!\in\!\{1,\dots,5\}\) (``hops'').  
Each ground-truth veracity vector \(v_{z}^{*}\in\{0,1\}^{|z|}\) is corrupted into \(v_{z}\) by independently negating every entry with probability \(0.5\).

\paragraph{\textsc{GSM8K}.}
We retain the original question–answer pairs \( (x,y,z)\) and obtain reference chains of thought \(z^{*}\) by querying \texttt{GPT-4.1} with $T=0.1$ under a deterministic JSON schema.
Corruption is applied by perturbing numerical constants in \(z^{*}\): each integer is randomly shifted by \(\pm1\), doubled, or halved so that exactly \(50\%\) of all statements become incorrect.

\paragraph{\textsc{CommonsenseQA}.}
Reasoning chains $z^{*}$ are generated similarly to \textsc{GSM8K}, by prompting \texttt{GPT-4.1} to produce structured explanations conditioned on the correct answer $y^*$. 
Corruption is applied as in \textsc{ProntoQA}, where each statement in $z^{*}$ is independently negated with probability $0.5$, yielding corrupted chains $z$ with associated veracity labels $v_z$.

\subsection{Baseline Inference Methods}

All baselines bypass the intractable problem of in-filling $V_z$ in the sequence  
\(X \!\rightarrow\! Z \!\rightarrow\! (V_{z}) \!\rightarrow\! Y\)  
by querying for the probability of \(v_{z}\) after observing \((x,y,z)\).
Unless stated otherwise, the sampling temperature is \(T\!=\!0.01\).

\begin{itemize}
    \item \textbf{Many2Many}: one-shot prediction of the full vector \(v_{z}\). 
    \item \textbf{Many2Many+CoT}: generates an intermediate rationale \(R_{V_z}\) before emitting \(v_{z}\).
    \item \textbf{Majority Voting}: draws \(M\!=\!50\) samples at \(T\!=\!0.5\) and returns the element-wise majority.
    \item \textbf{Recursive}: predicts labels sequentially, conditioning on past \(\langle z_i,v_{z_i}\rangle\) pairs.
\end{itemize}

We provide the prompt templates below:

\paragraph{Logical reasoning and commonsense reasoning.}
\begin{lstlisting}[language=text,frame=single,basicstyle=\ttfamily\small]
(... few shot demos with intruction)
### Context
{X}

### Query
{logical_question which is last sentense of X}

### Answer
{Y}

### Explanation Steps
Step 1: {Z_1}
...
Step N: {Z_N}

Give your judgement in JSON:
{"Label": [true, false, ...]}
\end{lstlisting}
\clearpage
\paragraph{Mathematical reasoning.}
\begin{lstlisting}[language=text,frame=single,basicstyle=\ttfamily\small]
(... few shot demos with intruction)
### Problem
{X}

### Answer
{Y}

### Solution Steps
Step 1: {Z_1}
...
Step N: {Z_N}

Give your judgement in JSON:
{"Label": [true/false, ...]}
\end{lstlisting}

\paragraph{Prompt Example (Logic, \textsc{ProntoQA}).}

This is a prompt example for Many2Many inference:

\begin{lstlisting}[language=text,frame=single,basicstyle=\ttfamily\scriptsize]
(... few shot demos with intruction)
### Context
Jompuses are overcast. Jompuses are yumpuses. Every yumpus is an impus.
Yumpuses are wooden. Lempuses are jompuses. Each impus is a gorpus.
Gorpuses are not transparent. Grimpuses are not sweet. Each impus is not nervous.
Jompuses are tumpuses. Zumpuses are brimpuses. Dumpuses are not dull.
Every lempus is a dumpus. Every numpus is not overcast. Each zumpus is orange.
Each impus is a shumpus. Every lempus is slow. Every tumpus is discordant.
Yumpuses are grimpuses. Polly is a jompus. Polly is a zumpus.

### Query
True or false: Polly is overcast.

### Answer
True

### Explanation Steps
Step 1: Polly is a jompus.
Step 2: Jompuses are overcast.
Step 3: Polly is overcast.

Give your judgement in JSON:
{"Label": [false, true, false]}
\end{lstlisting}

This is a prompt example for the recursive inference:

\begin{lstlisting}[frame=single,basicstyle=\ttfamily\scriptsize]
(... few shot demos with intruction)
### Context
Jompuses are overcast. Jompuses are yumpuses. Every yumpus is an impus.
Yumpuses are wooden. Lempuses are jompuses. Each impus is a gorpus.
Gorpuses are not transparent. Grimpuses are not sweet. Each impus is not nervous.
Jompuses are tumpuses. Zumpuses are brimpuses. Dumpuses are not dull.
Every lempus is a dumpus. Every numpus is not overcast. Each zumpus is orange.
Each impus is a shumpus. Every lempus is slow. Every tumpus is discordant.
Yumpuses are grimpuses. Polly is a jompus. Polly is a zumpus.

### Query
True or false: Polly is overcast.

### Answer
True


### Explanation Steps (with labels so far)
Step 1: Polly is a jompus.  Label: true (self verified)
...
Step k-1: Jompuses are overcast.  Label: false (self verified)
Step k:  Polly is overcast.  Label:

Predict the label for last only.
Return JSON: {"Label": true|false}
\end{lstlisting}

These prompts are recursively queried for $k=1,\ldots,N$. 
\clearpage
\paragraph{Prompt Example (Commonsense, \textsc{CommonsenseQA}).}

This is a prompt example for Many2Many inference:

\begin{lstlisting}[language=text,frame=single,basicstyle=\ttfamily\scriptsize]
(... few shot demos with instruction)
### Question
Where do you find wild cats?

### Query
A) trouble, B) dog's mouth, C) nature, D) floor, E) warm place.

### Answer
C

### Explanation Steps
Step 1: The question asks where wild cats are found.
Step 2: Option C is 'nature'.
Step 3: Wild cats are animals that live in natural environments.
Step 4: Nature refers to the outdoors and natural habitats.
Step 5: Therefore, Option C (nature) correctly answers where wild cats are found.

Give your judgement in JSON:
{"Label": [true, false, true, true, false]}
\end{lstlisting}

This is a prompt example for Recursive inference:

\begin{lstlisting}[frame=single,basicstyle=\ttfamily\scriptsize]
(... few shot demos with instruction)
### Question
Where do you find wild cats?

### Query
A) trouble, B) dog's mouth, C) nature, D) floor, E) warm place.

### Answer
C

### Explanation Steps (with labels so far)
Step 1: The question asks where wild cats are found. Label: true
Step 2: Option C is 'nature'. Label: false
Step 3: Wild cats are animals that live in natural environments. Label: false
Step 4: Nature refers to the outdoors and natural habitats. Label: true
Step 5: Therefore, Option C (nature) correctly answers where wild cats are found. Label:

Predict the label for last only.
Return JSON: {"Label": true|false}
\end{lstlisting}

These prompts are recursively queried for $k=1,\ldots,N$.

\clearpage
\paragraph{Prompt Example (Math, \textsc{GSM8K}).}

This is a prompt example for Many2Many inference:

\begin{lstlisting}[language=text,frame=single,basicstyle=\ttfamily\scriptsize]
(... few shot demos with intruction)
### Problem
Janet's ducks lay 16 eggs per day. She eats three for breakfast every morning
and bakes muffins for her friends every day with four. She sells the remainder
at the farmers' market daily for $2 per fresh duck egg. How much in dollars
does she make every day at the farmers' market?

### Answer
18

### Solution Steps
Step 1: Determine the total number of eggs laid by Janet's ducks each day.
        Intermediate output: 16
Step 2: Subtract the number of eggs Janet eats for breakfast from the total eggs.
        Intermediate output: 13
Step 3: Subtract the number of eggs used for baking muffins from the remaining eggs.
        Intermediate output: 9
Step 4: Identify the number of eggs Janet has left to sell at the farmers' market.
        Intermediate output: 9
Step 5: Determine the price per egg Janet sells at the market.
        Intermediate output: 2
Step 6: Multiply the number of eggs Janet sells by the price per egg to find her earnings.
        Intermediate output: 18
Step 7: Verify that 9 eggs multiplied by $2 per egg equals $18.
        Intermediate output: 18

Give your judgement in JSON:
{"Label": [false, false, false, true, false, true, true]}
\end{lstlisting}

This is a prompt example for Recursive inference:

\begin{lstlisting}[frame=single,basicstyle=\ttfamily\scriptsize]
(... few shot demos with intruction)
### Problem
Janet's ducks lay 16 eggs per day. She eats three for breakfast every morning
and bakes muffins for her friends every day with four. She sells the remainder
at the farmers' market daily for $2 per fresh duck egg. How much in dollars
does she make every day at the farmers' market?

### Answer
18


### Solution Steps (with labels so far)
Step 1: Determine the total number of eggs laid by Janet's ducks each day.
        Intermediate output: 16. Label: true
...
Step k: Identify the number of eggs Janet has left to sell at the farmers' market.
        Intermediate output: 9. Label: true
Step k+1: Determine the price per egg Janet sells at the market.
        Intermediate output: 2. Label:

Predict the label for last only.
Return JSON: {"Label": true|false}
\end{lstlisting}

\clearpage
\subsection{Veracity Search (VS)}

VS keeps the original trajectory
\(X \!\rightarrow\! Z \!\rightarrow\! V_{z} \!\rightarrow\! Y\)
and optimises \(V_{z}\) with a Metropolis algorithm that leverages the latent variable model-based posterior
\(
R(V_{z}) = \PLM(Y, V_{z}\mid X,Z)
\)
as a reward. In this section, we describe the detailed setting and prompt of our method.

\begin{description}[leftmargin=1.8em]
\item[Iterations] 200
\item[Temperature] linear annealing from \(T_0\!=\!2.0\) to \(T_{200}\!=\!1.0\)
\item[Proposal] single-bit flip (Hamming-1)
\end{description}

\paragraph{Reward Prompt (Logic and Commonsense).}
\begin{lstlisting}[language=text,frame=single,basicstyle=\ttfamily\small]
(... few shot demos with instruction)
Label **True** if the step follows from Context, **False** otherwise.

### Context
{X}

### Query
{logical_question (last sentense of X)}

### Explanation Steps
{Z_1 ... Z_N}

### Label Vector (V_z)
{0/1 sequence}

### Answer
{Y}
\end{lstlisting}

\paragraph{Reward Prompt (Math).}
\begin{lstlisting}[language=text,frame=single,basicstyle=\ttfamily\small]
(... few shot demos with instruction)
Think step by step. Mark each step as **Correct** or **Incorrect**.

### Question
{X}

### Solution Steps
{Z_1 ... Z_N}

### Answer
{Y}
\end{lstlisting}

\clearpage
\subsection{Amortized Veracity Inference (AVI)}

\paragraph{Training.} We amortize VS into a lightweight veracity inference machine by fine-tuning Qwen3-4B/8B on 5,000 labeled contexts produced by the VS.

\begin{itemize}
    \item \textbf{LoRA} rank 8, \(\alpha\!=\!32\)
    \item \textbf{Batch / Accum.} 32 / 8
    \item \textbf{Optimizer / LR} AdamW~\citep{loshchilov2017decoupled}, \(1\!\times\!10^{-4}\)
    \item \textbf{Hardware} single NVIDIA A100L (80GB)
    \item \textbf{Runtime} 16 min (4B) / 24 min (8B) per epoch
\end{itemize}

\begin{figure}[htb]
  \centering          
  \includegraphics[width=0.4\linewidth]{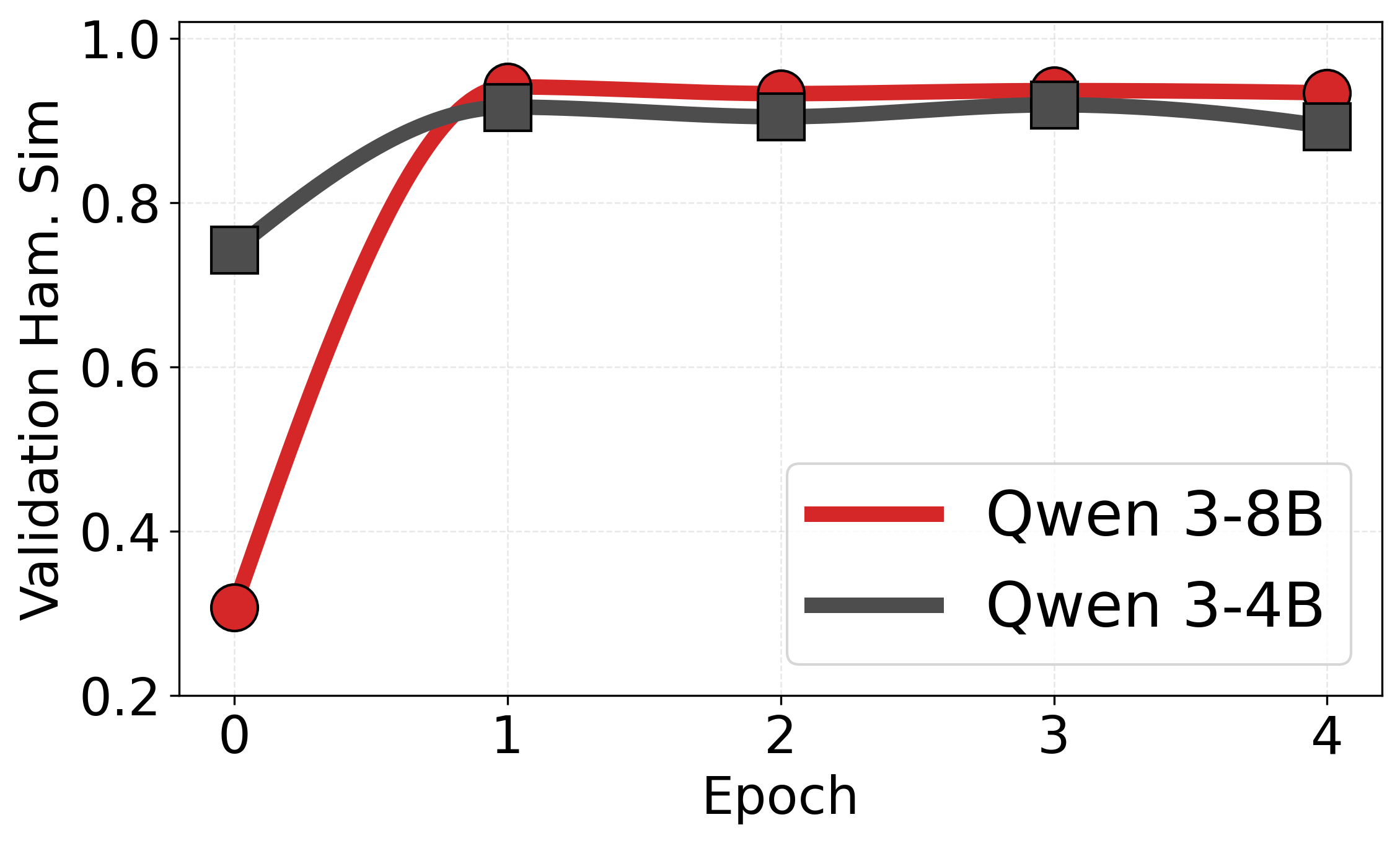}
  \caption{Validation curve.}\label{fig:validation}
\end{figure}

Validation accuracy saturates after a single epoch (Fig.~\ref{fig:validation}).

\paragraph{Inference for $V_z$.} We do Many2Many inference without accessing true answer $Y$. The prompt template is as follows:
\begin{lstlisting}[language=text,frame=single,basicstyle=\ttfamily\small]
(... few shot demos with intruction)
### Context
{X}

### Query
{logical_question which is last sentense of X}

### Answer
Unknown

### Explanation Steps
Step 1: {Z_1}
...
Step N: {Z_N}

Give your judgement in JSON:
{"Label": [true, false, ...]}
\end{lstlisting}

\paragraph{Inference for $Y$. } Using corrected $Z$ with $V_z$ by negating for False statement, we can infer $Y$. The prompt template is as follows:

\begin{lstlisting}[language=text,frame=single,basicstyle=\ttfamily\small]
(... few shot demos with intruction)
### Context
{X}

### Query
{logical_question which is last sentense of X}

### Explanation Steps
Step 1: {Z_1}
...
Step N: {Z_N}

Your answer:
\end{lstlisting}

\clearpage

\section{Natural language examples}

We provide 3 examples (first 3 examples in a batch) for logical and math reasoning correction using VS.

\paragraph{Logic example 1: Max is not wooden.}

\begin{lstlisting}
### Context (X)
Numpuses are not wooden. Vumpuses are lempuses. Rompuses are not dull. ...
Max is a vumpus. Max is a lorpus.

### Query (X)
True or false: Max is not wooden.

### Answer (Y)
True

### Corrupted chain (Z)
1. Max is not a vumpus.
2. Each vumpus is not a brimpus.
3. Max is a brimpus.
4. not Brimpuses are grimpuses.
5. Max is not a grimpus.
6. Grimpuses are shumpuses.
7. Max is a shumpus.
8. Shumpuses are numpuses.
9. Max is a numpus.
10. Numpuses are wooden.
11. Max is wooden.

### V_z pred
[F, F, T, F, F, T, T, T, T, T, T]

### V_z ground truth
[F, F, T, F, F, T, T, T, T, F, F]
\end{lstlisting}

\clearpage
\paragraph{Logic example 2: Fae is not discordant.}

\begin{lstlisting}
### Context (X)
Every rompus is not sweet. ... Fae is a sterpus. Fae is a brimpus.

### Query (X)
True or false: Fae is not discordant.

### Answer (Y)
False

### Corrupted chain (Z)
1. Fae is a brimpus.
2. Brimpuses are rompuses.
3. Fae is not a rompus.
4. Rompuses are lempuses.
5. Fae is not a lempus.
6. Each lempus is not a grimpus.
7. Fae is not a grimpus.
8. Each grimpus is a lorpus.
9. Fae is a lorpus.
10. Each lorpus is not discordant.
11. Fae is not discordant.

### V_z pred
[T, T, F, T, F, F, F, T, T, F, F]

### V_z ground truth
[T, T, F, T, F, F, F, T, T, F, F]
\end{lstlisting}

\paragraph{Logic example 3: Stella is liquid.}

\begin{lstlisting}
### Context (X)
Every sterpus is not melodic. ... Stella is a rompus. Stella is a dumpus.

### Query (X)
True or false: Stella is liquid.

### Answer (Y)
True

### Corrupted chain (Z)
1. Stella is a dumpus.
2. Each dumpus is a shumpus.
3. Stella is not a shumpus.
4. not Shumpuses are sterpuses.
5. Stella is a sterpus.
6. not Sterpuses are gorpuses.
7. Stella is not a gorpus.
8. Each gorpus is a jompus.
9. Stella is a jompus.
10. not Jompuses are liquid.
11. Stella is not liquid.

### V_pred
[T, T, F, F, F, F, T, T, T, F, F]

### V_z ground truth
[T, T, F, F, T, F, F, T, T, F, F]
\end{lstlisting}

\clearpage

\paragraph{Commonsense example 1: Salad.}

\begin{lstlisting}
### Question (X)
What meal usually has carrots and mostly leafy vegetables?

### Choices (Y)
A) cellar, B) salad, C) store, D) refrigerator, E) soup.

### Answer (T_Y)
B

### Corrupted chain (Z)
1. The question asks for a meal that usually has not carrots and mostly leafy vegetables.
2. Option B is not 'salad'.
3. A salad is not a meal that typically consists of mostly leafy vegetables.
4. Carrots are not commonly included as an ingredient in salads.
5. Therefore, Option B (salad) fits the description given in the question.

### V_z pred
[T, F, F, F, T]

### V_z ground truth
[F, F, F, F, T]
\end{lstlisting}

\paragraph{Commonsense example 2: Temperate climate.}

\begin{lstlisting}
### Question (X)
What is the easiest place for a human or another animal to live?

### Choices (Y)
A) workplace, B) gravesite, C) shelter, D) interesting place, E) temperate climate.

### Answer (T_Y)
E

### Corrupted chain (Z)
1. Not true: The question asks for the easiest place for a human or another animal to live.
2. Option E is 'temperate climate'.
3. False: A temperate climate provides moderate temperatures and conditions suitable for most humans and animals.
4. Other options such as workplace, gravesite, shelter, or interesting place do not inherently provide suitable living conditions for both humans and animals.
5. Therefore, Option E (temperate climate) is not the correct answer.

### V_z pred
[F, T, F, T, F]

### V_z ground truth
[F, T, F, T, F]
\end{lstlisting}

\clearpage
\paragraph{Commonsense example 3: Virginia.}

\begin{lstlisting}
### Question (X)
Which state is the Blue Ridge Mountains Roanoke City in?

### Choices (Y)
A) appalachians, B) north carolina, C) virginia, D) pennsylvania, E) montana.

### Answer (T_Y)
C

### Corrupted chain (Z)
1. The question asks which state the Blue Ridge Mountains Roanoke City is not in.
2. Option C is not 'virginia'.
3. Roanoke City is not located in the state of Virginia.
4. Not true: The Blue Ridge Mountains extend through Virginia.
5. Incorrect: Therefore, Option C (virginia) correctly answers the question.

### V_z pred
[F, F, F, F, F]

### V_z ground truth
[F, F, F, F, F]
\end{lstlisting}

\clearpage
\paragraph{Math example 1: Janet's duck-egg earnings.}

\begin{adjustbox}{max width=\linewidth}
\begin{lstlisting}
### Problem (X)
Janet's ducks lay 16 eggs per day. She eats three ... How much in dollars
does she make every day at the farmers' market?

### Answer (Y)
18

### Corrupted chain (Z)
1. Determine total eggs laid.  Out: 17
2. Subtract breakfast eggs.    Out: 12
3. Subtract muffin eggs.       Out: 10
4. Eggs left to sell.          Out: 9
5. Price per egg.              Out: 1
6. Multiply eggs x price.      Out: 18
7. Verify 9 x $2 = $18.        Out: 18

### V_z pred
[F, F, F, F, F, F, T]

### V_z ground truth
[F, F, F, T, F, T, T]
\end{lstlisting}
\end{adjustbox}

\paragraph{Math example 2: Robe fabric bolts.}

\begin{adjustbox}{max width=\linewidth}
\begin{lstlisting}
### Problem
A robe takes 2 bolts of blue fiber and half that much white fiber.
How many bolts in total does it take?

### Answer
3

### Corrupted chain (Z)
1. Identify blue bolts: 2  Out: 1
2. White bolts = half.      (no out)
3. Half of 2 is 1 bolt.      Out: 1
4. Add 2 + 1.               Out: 3
5. Verify total = 3.        Out: 4

### V_z pred
[T, T, T, T, F]

### V_z ground truth
[F, T, T, T, F]
\end{lstlisting}
\end{adjustbox}

\clearpage
\paragraph{Math example 3: House-flipping profit.}

\begin{adjustbox}{max width=\linewidth}
\begin{lstlisting}
### Problem
Josh buys a house for $80,000 and spends $50,000 on repairs,
which increase the value by 150%. How much profit did he make?

### Answer
70000

### Corrupted chain (Z)
1. Purchase price.                      Out: 80001
2. Repair cost.                         Out: 50000
3. Total investment.                    Out: 130000
4. Repairs increase value by 150%.      (no out)
5. 150% of 80k.                         Out: 119999
6. New value after repairs.             Out: 200000
7. (Erroneous) value double-count.      Out: 400000
8. Profit = 200k - 130k.                Out: 70000
9. Verification step.                   Out: 140000

### V_pred
[T, T, T, T, F, T, F, T, F]

### V_z ground truth
[F, T, T, T, F, T, F, T, F]
\end{lstlisting}
\end{adjustbox}

\clearpage

\section{Additional results}

\subsection{Analysis of Robustness to Different Error Distributions}
\label{app:different_patterns}

Reasoning errors exhibit complex dependencies across reasoning steps, and won't necessarily be distributed uniformly. We therefore evaluate our method under different error distributions using both synthetic CoT corruption methods (\S\ref{app:synthetic_error_patterns}) and naturally-occuring errors (\S\ref{app:cot_wild}).

\subsubsection{Different Patterns of Synthetic CoT Corruption}
\label{app:synthetic_error_patterns}
Mistakes can concentrate near the beginning of a CoT (e.g. when premises are misapplied), or toward the end (e.g. when incorrect conclusions are inferred from preceding assertions). To examine robustness under such conditions, we evaluate three perturbation patterns in \textsc{ProntoQA}: \emph{front-side} (errors injected in the first half of reasoning steps through negating correct statements), \emph{uniform}, and \emph{back-side} (second half of statements are negated). Results obtained with Qwen3-8B are shown in Table~\ref{tab:false-distr}.

\begin{table}[h]
  \centering
  \small
  \caption{Average Hamming Similarity and Exact-Match Accuracy across 100 samples of 5-hop \textsc{ProntoQA} using Qwen3-8B under different error patterns.}
  \label{tab:false-distr}
  \begin{tabular}{lcccccc}
    \toprule
    & \multicolumn{3}{c}{Hamming Similarity} & \multicolumn{3}{c}{Exact Match} \\
    \cmidrule(lr){2-4}\cmidrule(lr){5-7}
    Method & Front-side & Uniform & Back-side & Front-side & Uniform & Back-side \\
    \midrule
    Recursive        & 0.768 & 0.679 & 0.707 & 0.260 & 0.030 & 0.030 \\
    Many2Many        & 0.857 & 0.678 & 0.684 & 0.000 & 0.010 & 0.000 \\
    \textbf{VS}      & \textbf{0.983} & \textbf{0.963} & \textbf{0.947} & \textbf{0.810} & \textbf{0.720} & \textbf{0.750} \\
    \bottomrule
  \end{tabular}
\end{table}

Across all perturbation patterns, VS consistently surpasses both Recursive and Many2Many baselines. In particular, it achieves near-perfect Hamming similarity (0.947–0.983) and markedly higher exact-match accuracy (0.720–0.810), whereas baseline accuracies remain close to zero. These findings indicate that VS is robust to different distributions of errors within the reasoning chain, underscoring its applicability to realistic settings where error patterns are diverse and not easily predictable.

\subsubsection{Naturally Occurring (LM-Generated) Errors}
\label{app:cot_wild}

The main results presented in this paper are derived from controlled experiments using synthetically corrupted reasoning chains, 
facilitating the evaluation of the accuracy of the inferred veracity assignments $v_z$. However, a practical scenario involves an LM generating a  CoT $z$ on its own, with errors arising naturally as a result of the limitations of the model's reasoning capabilities. Here, we examine whether our method generalizes to this more realistic setting. 

Reasoning chains $z$ were generated using structured-decoding with a Qwen3-4B model for questions $x$ sampled from \textsc{ProntoQA} and \textsc{CommonsenseQA} datasets ($5,000$ samples each). The resulting $(x,z)$ pairs and corresponding answer labels $y^*$ were then subjected to \textsc{VS} to obtain predicted veracities $\hat{v}_z \in \{\texttt{True},\texttt{False}\}^{|z|}$.

\begin{table}[htb!]
  \centering
  \small
  \caption{Veracity accuracy in naturally generated CoTs (1,000 test samples for each task).}
  \label{tab:natural_vz_accuracy}
  \begin{tabular}{lcccc}
    \toprule
    & \multicolumn{2}{c}{Hamming Similarity} & \multicolumn{2}{c}{Exact Match Accuracy} \\
    \cmidrule(lr){2-3}\cmidrule(lr){4-5}
    Method & ProntoQA & CommonsenseQA & ProntoQA & CommonsenseQA  \\
    \midrule
    Many2Many        & 0.74 & 0.59  & 0.00 & 0.00\\
    VS      & \textbf{0.92} & \textbf{0.86} & \textbf{0.50} & \textbf{0.59} \\
    \bottomrule
  \end{tabular}
\end{table}

We evaluated VS on 1,000 test samples (for each dataset) in terms of veracity inference accuracy (Hamming similarity and exact match accuracy), by comparing predicted $\hat v_z$ to pseudo-ground-truth labels $v_z^*$ obtained using a GPT-4.1 oracle. The results shown in Table~\ref{tab:natural_vz_accuracy} suggest that VS maintains a strong advantage over the in-context learning baseline (Many2Many) under this natural error distribution.

\subsection{Self-Correction/Improvement of LM-Generated Reasoning Chains}
\label{app:self-refine}

Following from the experiment detailed in \S\ref{app:cot_wild}, where VS was applied to naturally-occurring errors, we then sought to evaluate whether AVI can be used as a feedback signal as part of a self-correction/improvement framework for improving reasoning performance. Concretely, we fine-tuned a Qwen3-4B model using AVI on predicted veracities sampled using VS for $10,000$ $(x,z,y^*)$ triples, to learn a distribution $Q(v_z \mid x,z)$ that does not depend on $y^*$. The training setup is otherwise identical to the one described in \S\ref{sec:amortized_corrector}.
Then, we used AVI to predict erroneous steps in $z$ by inferring stepwise veracities $\hat{v}_z \sim Q(v_z \mid x,z)$.

\begin{table}[t]
  \centering
  \small
  \caption{Answer accuracy on 1000 \textsc{ProntoQA} problems. Standard deviation is reported on five intendant runs.}
  \label{tab:prontoqa_wild}
  \begin{tabular}{lc}
    \toprule
    Correction Strategy & Accuracy $\uparrow$ \\
    \midrule
    No Correction (Raw CoT) &$0.712 \pm 0.002$ \\
    Correction using Many2Many & $0.697 \pm0.008$ \\
    Correction using AVI & \textbf{0.730 $\pm$ 0.002
} \\
    \bottomrule
  \end{tabular}
\end{table}

In the case of \textsc{ProntoQA}, we corrected steps predicted to be \texttt{False} by negating them, and queried a Qwen3-4B model to predict the answer $y$ conditioned on the corrected CoT. 

Table~\ref{tab:prontoqa_wild} shows that standard self-correction using in-context learning (Many2Many) to provide feedback via veracity assignments does not improve performance in \textsc{ProntoQA}: the accuracy even drops from $0.712$ to $0.697$. This highlights that applying correction without reliable veracity identification may in fact be harmful. In contrast, our AVI-based correction increases accuracy to $0.730$, showing that once veracity is accurately identified, correction becomes beneficial rather than detrimental. Without accurate veracity signals, correction may propagate or even amplify errors.

Next, we moved beyond simple correction via negation by testing the applicability of our approach in a more realistic setting where negation-based edits aren't a natural way to correct. In particular, we consider \textsc{CommonsenseQA}, where reasoning chains involve open-ended statements in natural language, and correction is more easily conducted as part of inference-time self-improvement methods such as \emph{Self-Refine}~\citep{madaan2023self}, which iteratively alternates between a \emph{feedback model} and a \emph{refinement model}. We adapt this framework by replacing the baseline feedback model (which is a few-shot prompted LM, similar to our Many2Many baseline) with our AVI machine. Given a CoT $z$ generated by Qwen3-4B, the AVI model first predicts veracity labels $v_z$ that identify incorrect reasoning steps. The refinement model (a few-shot-prompted Qwen3-4B model) then resamples downstream reasoning steps, conditioned on these veracity labels. This allows AVI to supply explicit supervision on \emph{what to correct}, while the refinement model handles the problem of \emph{how to correct}. 

\begin{table}[t]
  \centering
  \small
  \caption{Reasoning accuracy on \textsc{CommonsenseQA} with one iteration of Self-Refine. Accuracy is averaged over 5 random seeds ($\pm$ standard deviation).}
  \label{tab:commonsense_selfrefine}
  \begin{tabular}{lc}
    \toprule
    Method & Accuracy $\uparrow$ \\
    \midrule
    Original Reasoning & $0.741 \pm 0.001$ \\
    Self-Refine (Many2Many) & $0.749 \pm 0.005$ \\
    Self-Refine (AVI) & \textbf{0.756 $\pm$ 0.002} \\
    \bottomrule
  \end{tabular}
\end{table}

The results are shown in Table~\ref{tab:commonsense_selfrefine} for 1{,}000 test questions from \textsc{CommonsenseQA}. 
The gain from integrating AVI into Self-Refine is nearly twice as large as the improvement obtained with a standard Many2Many feedback model ($+0.008$), i.e.\ $0.015/0.008 \approx 1.9\times$, though in  absolute terms may appear modest ($+0.015$ over the raw baseline). 
In general, the trend observed in all our experiments is that our method improves the accuracy of error-identification (whether using VS or AVI) by a larger margin over baselines than it improves reasoning accuracy post-correction, hinting at a bottleneck in reasoning capabilities in the underlying model. This reflects the fact that verification may be easier than generation, and that more work is needed to find ways to make better use of the more robust veracity signal provided by VS and AVI in downstream reasoning tasks. 

\subsection{Block Metropolis}
\label{app:block}

Our default implementation of VS uses single-bit Metropolis updates, where one veracity label is flipped at a time. While simple and effective, this approach may be less effective if errors are correlated, requiring simultaneous updates to transition between distinct high-reward modes. To examine this possibility, we extended our sampler with \emph{block Metropolis} updates, in which random contiguous blocks of size 1, 2, or 3 are flipped together.

\begin{table}[h]
  \centering
  \small
  \caption{Average Hamming similarity of predicted veracity vectors using single-bit vs.\ block Metropolis updates on 1{,}000 samples with Qwen3-4B.}
  \label{tab:block_metropolis}
  \begin{tabular}{lccc}
    \toprule
    Method & \textsc{ProntoQA} & \textsc{GSM8K} & \textsc{CommonsenseQA} \\
    \midrule
    Single-bit Metropolis & 0.910 & 0.711 & 0.860 \\
    Blocked Metropolis    & 0.918 & 0.704 & 0.867 \\
    \bottomrule
  \end{tabular}
\end{table}

As shown in Table~\ref{tab:block_metropolis}, both variants perform comparably across the three tasks. This suggests that single-bit updates are already sufficient for short to medium-length reasoning chains in the domains of reasoning that we tested. Nonetheless, block updates represent a natural extension of our framework, with potential advantages for longer sequences and settings where LMs produce errors with more complex inter-dependencies.

Our approach is compatible with other proposal strategies, such as adaptive proposals, or gradient-informed updates, making them promising directions for future exploration.

\subsection{Extending to Categorical Variables}
\label{app:categorical}

In many contexts, binary veracity labels may be too restrictive. For example, when verifying properties such as the \emph{relevance} of a reasoning step, or when allowing for ambiguity in correctness, it can be useful for the latent variable $V_z$ to take on more than two values. While this enlarges the label space, it remains far smaller than the full space of natural language sequences, making inference tractable.

Our framework naturally generalizes to this setting by replacing the binary vector with a $k$-class categorical vector and augmenting the MCMC transition table accordingly. To illustrate, we extended \textsc{ProntoQA} with a third label, ``Unrelated,'' by injecting unrelated---but factually correct---statements into the reasoning chain from a small predefined pool (e.g., ``Humans are animals.'' or ``The sky is blue.''). 

\begin{table}[h]
  \centering
  \small
  \caption{Veracity prediction on \textsc{ProntoQA} with three classes: True, False, and Unrelated. Average Hamming similarity across 100 samples is reported.}
  \label{tab:categorical_prontoqa}
  \begin{tabular}{lc}
    \toprule
    Method & Hamming Similarity $\uparrow$ \\
    \midrule
    Many2Many & $0.66$ \\
    VS & $\mathbf{0.91}$ \\
    \bottomrule
  \end{tabular}
\end{table}

These preliminary results suggest that our approach extends naturally beyond binary veracity, and can accommodate richer categorical variables for reasoning verification.

\subsection{Correlation Analysis}
\label{app:correlation}

Our working hypothesis (see Section~\ref{sec:method}) is that veracity assignments $v_z$ which maximize the joint-likelihood proxy reward 
\[
\PLM(v_z\, y^* \mid x, z) \;\propto\; \mathbb{P}(V_z = v_z \mid Y = y^*, x, z)
\] 
will tend to be closer to the ground-truth veracity $v_z^*$. Intuitively, if the accuracy of a language model’s final prediction depends on the correctness of its reasoning steps, then higher joint likelihood should correlate with more accurate veracity labels. 

To validate this hypothesis, we measured the correlation between the joint likelihood and veracity accuracy. For each problem we enumerated all possible veracity vectors ($2^N$ assignments for $N{=}7$ reasoning steps) and computed both their joint likelihood $\PLM(v_z\, y^* \mid x, z)$ and their Hamming similarity with the ground-truth vector $v_z^*$. Table~\ref{tab:correlation} reports the average Pearson correlation across 100 randomly selected problems drawn from \textsc{ProntoQA} and \textsc{CommonsenseQA} datasets.

\begin{table}[t]
  \centering
  \small
  \caption{Pearson correlation between Hamming similarity and joint likelihood across all possible veracity assignments (7 statements per sample). Results are averaged over 100 samples for each dataset.}
  \label{tab:correlation}
  \begin{tabular}{lcc}
    \toprule
    Model & \textsc{ProntoQA} & \textsc{CommonsenseQA} \\
    \midrule
    Qwen3-4B & 0.56 & 0.72 \\
    Qwen3-8B & 0.74 & 0.78 \\
    LLaMA 3B & 0.72 & 0.67 \\
    LLaMA 8B & 0.70 & 0.79 \\
    \bottomrule
  \end{tabular}
\end{table}

These results suggest that veracity vectors with higher joint likelihood also achieve higher Hamming similarity with the ground truth. In other words, the proxy reward used in our search procedure is not only theoretically motivated but also empirically aligned with veracity accuracy. This correlation provides a clear post-hoc explanation for why our method consistently yielded strong Hamming similarity throughout the main experiments, thereby grounding the observed performance gains in a measurable property of language models.

\subsection{The Cost of Veracity Search on Inference-Time}
\label{app:complexity}

\begin{table}[t]
  \centering
  \small
  \caption{Average VS wall-time over 5 independent runs as a function of CoT length.}
  \label{tab:walltime}
  \begin{tabular}{lc}
    \toprule
    Number of Reasoning Steps & Time (sec) \\
    \midrule
    3 & 7.33  \\
    5 & 7.83  \\
    7 & 11.14  \\
    9 & 12.19  \\
    11 & 13.09  \\
    \bottomrule
  \end{tabular}
\end{table}

In VS, the CoT $z$ is fixed, and we score a proposed veracity assignment $v_z$ via the joint likelihood $P_{\mathrm{LM}}(v_z, y^{*} \mid x, z)$, where $y^{*}$ is the correct answer to the question in $x$. This likelihood is computed with teacher‑forcing: the concatenated sequence $(v_z, y^{*})$ is processed in a single forward pass (prefill stage), and logits for all tokens are produced under causal self‑attention. This avoids the substantially more expensive sequential autoregressive decoding required by in‑context learning baselines (Many2Many). However, because VS typically requires scoring multiple proposals as part of the search procedure, this advantage diminishes as the number of VS iterations increases. Empirically, we find that our method provides a favorable trade‑off between inference time and veracity accuracy, as shown in Figure~\ref{fig:inference-eff}, but we provide a complementary analysis of the computational cost incurred from VS iterations below.

Let $L_\text{context} = |(x, z)|$ and $L_\text{tail} = |(v_z, y^*)|$. Naively, across $N$ VS iterations, the cost scales as $O\big(N (L_\text{context} + L_\text{tail})^2\big)$ due to the quadratic cost of self-attention. We can reduce computational requirements with prefix key‑value caching for the fixed context $(x, z)$, so the overall cost becomes $O\big(L_\text{context}^2\big) + O\big(N (L_\text{context}\, L_\text{tail} + L_\text{tail}^2)\big)$. For proposals that differ from a reference $v_z$ at a single position $i$, we can also cache the extended prefix $(x, z, v_z[:i])$. If $i$ is uniformly distributed, the expected suffix length is $L_\text{tail}/2$, yielding a constant‑factor reduction: $O\big(L_\text{context}^2\big) + O\big(N (L_\text{context}\, L_\text{tail}/2 + L_\text{tail}^2/4)\big)$. Finally, batching proposals that share the same cached prefix (or the same divergence index $i$) lets the GPU process many tails in parallel, improving utilization and reducing wall‑clock time. In other words, we can trade memory for faster runtime, if desired.

We empirically evaluate how wall‑clock time varies as a function of CoT length using Qwen3‑4B-based VS in \textsc{ProntoQA}, for CoTs with lengths ranging from 3 to 11 steps. A single GPU (RTX A6000 ADA) was used, and VS involved 100 Metropolis iterations. The measurements displayed in Table~\ref{tab:walltime} reflect end‑to‑end VS, including greedy initialization and simulated annealing.

\subsection{Veracity Search on Smaller and Larger Models}
\label{app:model_size}
The original Veracity Search (VS) experiments were performed on 3B–8B models.
To examine model-size sensitivity, we additionally evaluated a smaller model (Qwen3-1.7B) and a larger model (Qwen3-14B).
The experimental design and search hyperparameters match those used to produce the results in Table~\ref{tab:combined_scores}.

\begin{table}[h]
\centering
\caption{Veracity inference accuracy (Hamming similarity) across three benchmarks (average over 1000 samples each).}
\label{tab:vs_modelsize}
\begin{tabular}{lcccccc}
\toprule
  \centering
\small
& \multicolumn{2}{c}{GSM8K} 
& \multicolumn{2}{c}{ProntoQA} 
& \multicolumn{2}{c}{CommonsenseQA} \\
\cmidrule(lr){2-3} \cmidrule(lr){4-5} \cmidrule(lr){6-7}
& 1.7B & 14B & 1.7B & 14B & 1.7B & 14B \\
\midrule
Many2Many & 0.512 & 0.675 & 0.529 & 0.714 & 0.488 & 0.524 \\
VS         & \textbf{0.657} & \textbf{0.833} & \textbf{0.713} & \textbf{0.928} & \textbf{0.832} & \textbf{0.980} \\
\bottomrule
\end{tabular}
\end{table}

Across all three tasks, larger models provide more accurate veracity predictions when using either VS or the in-context learning baseline (Many2Many). VS maintains a consistent advantage over this baseline, indicating that the joint-likelihood used to guide VS becomes increasingly informative as model capacity increases.

\section*{Large Language Model Usage}

Large language models (LLM) were used only for minor polishing of the writing quality both in the main text and in the code. They were also used to assist with debugging. Any modification to the text that was suggested by an LLM to improve clarity was verified by the authors. LLMs were not used to generate the ideas, methods, experimental designs, and analyses of results that are presented in this paper.

\end{document}